%% file: main.tex
\pdfoutput=1
\PassOptionsToPackage{dvipsnames}{xcolor}

\documentclass[11pt]{article}

\usepackage[final]{acl}

\usepackage{times}
\usepackage{latexsym}

\usepackage{xspace,mfirstuc,tabulary}
\usepackage{booktabs}
\usepackage{amssymb}
\usepackage{amsmath}
\usepackage{multirow,booktabs, hhline}
\usepackage[ruled,noend]{algorithm2e}
\usepackage{amsmath, bm}
\usepackage{graphicx}

\usepackage{cleveref}
\crefname{section}{§}{§§}
\Crefname{section}{§}{§§}

\usepackage{makecell}
\usepackage{adjustbox}
\usepackage{xcolor}
\usepackage{color, colortbl}
\definecolor{BLUE_D1}{HTML}{a9cce3}
\definecolor{GREEN_D1}{HTML}{b6d7a8}
\definecolor{ORANGE_D1}{HTML}{f5cba7}
\definecolor{GRAY_D1}{HTML}{d5dbdb}
\usepackage{graphicx}

\newcommand\RoBERTaBASE{RoBERTa$_{\textsc{Base}}$\xspace}
\newcommand\RoBERTaLARGE{RoBERTa$_{\textsc{Large}}$\xspace}
\newcommand\BertBASE{BERT$_{\textsc{Base}}$\xspace}
\newcommand\TSMALL{T5$_{\textsc{Small}}$\xspace}
\newcommand\TBASE{T5$_{\textsc{Base}}$\xspace}

\newcommand\TXXL{T5$_{\textsc{XXL}}$\xspace}
\newcommand\TPTtask{TPT$_{\textsc{task}}$\xspace}
\newcommand\TPTmodel{TPT$_{\textsc{model}}$\xspace}

\usepackage{bbm}
\usepackage{float}
\usepackage{subfigure}
\usepackage{warpcol}
\usepackage{makecell}
\usepackage{hyperref}

\usepackage{dcolumn,booktabs}
\newcolumntype{d}[1]{D{.}{.}{#1}}
\newcommand\mc[1]{\multicolumn{1}{c}{#1}} 
\newcommand\mcl[1]{\multicolumn{1}{c|}{#1}}
\usepackage[T1]{fontenc}
\usepackage[utf8]{inputenc}

\usepackage{microtype}

\title{On Transferability of Prompt Tuning for Natural Language Processing}

\author{
 Yusheng~Su$^{1,3}\thanks{\quad The first two authors contributed equally.}$\hspace{0.5em}, Xiaozhi~Wang$^{1,3*}$, Yujia~Qin$^{1,3}$, Chi-Min~Chan$^{1,3}$, Yankai~Lin$^4$, \\  \textbf{Huadong~Wang$^{1,3}$,
Kaiyue~Wen$^{1,3}$, Zhiyuan~Liu$^{1,3}\thanks{\quad Corresponding author: Z.Liu and M.Sun.}$\hspace{0.5em}, Peng~Li$^2$\thanks{\quad Partly done while P.Li was working at Tencent.}, Juanzi~Li$^{1,3}$}, \\ \textbf{Lei~Hou$^{1,3}$, Maosong~Sun$^{1,3+}$, Jie~Zhou$^4$} \\
 $^1$Department of Computer Science and Technology, BNRist;\\
 $^2$Institute for AI Industry Research (AIR);\\
  $^3$Institute for Artificial Intelligence, Tsinghua University, Beijing, 100084, China \\
 $^4$Pattern Recognition Center, WeChat AI, Tencent Inc, China. \\

    \tt{\{suys19,wangxz20\}@mails.tsinghua.edu.cn}\\
}

\begin{document}
\maketitle

\input{sections/abstract}
\input{sections/introduction}

\input{sections/related_work}

\input{sections/background}

\input{sections/cross_task}

\input{sections/cross_model}

\input{sections/similarity}

\input{sections/conclusion}

\input{sections/acknowledgments}
%\input{sections/appendix}

\bibliography{anthology,custom_output}
\bibliographystyle{acl_natbib}
\input{sections/appendix}

\end{document}

%% file: sections/abstract.tex
\begin{abstract}

Prompt tuning (PT) is a promising parameter-efficient method to utilize extremely large pre-trained language models (PLMs), which can achieve comparable performance to full-parameter fine-tuning by only tuning a few soft prompts. However, PT requires much more training time than fine-tuning. Intuitively, knowledge transfer can help to improve the efficiency. To explore whether we can improve PT via prompt transfer, we empirically investigate the transferability of soft prompts across different downstream tasks and PLMs in this work. We find that (1) in zero-shot setting, trained soft prompts can effectively transfer to similar tasks on the same PLM and also to other PLMs with a cross-model projector trained on similar tasks; (2) when used as initialization, trained soft prompts of similar tasks and projected prompts of other PLMs can significantly accelerate training and also improve the performance of PT. 
Moreover, to explore what decides prompt transferability, we investigate various transferability indicators and find that the overlapping rate of activated neurons strongly reflects the transferability, which suggests how the prompts \textit{stimulate} PLMs is essential. Our findings show that prompt transfer is promising for improving PT, and further research shall focus more on prompts' stimulation to PLMs. The source code can be obtained from \url{https://github.com/thunlp/Prompt-Transferability}.

\end{abstract}

%% file: sections/introduction.tex
\section{Introduction}
\label{label:introduction}

Pre-trained language models (PLMs), such as BERT~\citep{devlin-etal-2019-bert} and GPT~\citep{radford2018improving} have achieved great performance on various natural language processing (NLP) tasks~\citep{han2021pre}. Recently, people have found that extremely large PLMs can achieve remarkable improvements, and various large PLMs are continually developed~\citep{brown2020GPT3,raffel2020exploring,zhang2021cpm,zeng2021pangu,wei2021finetuned,Sun2021ERNIE3L}, which contain up to hundreds of billions of parameters.

\begin{figure}[!t]
\centering
\includegraphics[width=\columnwidth]{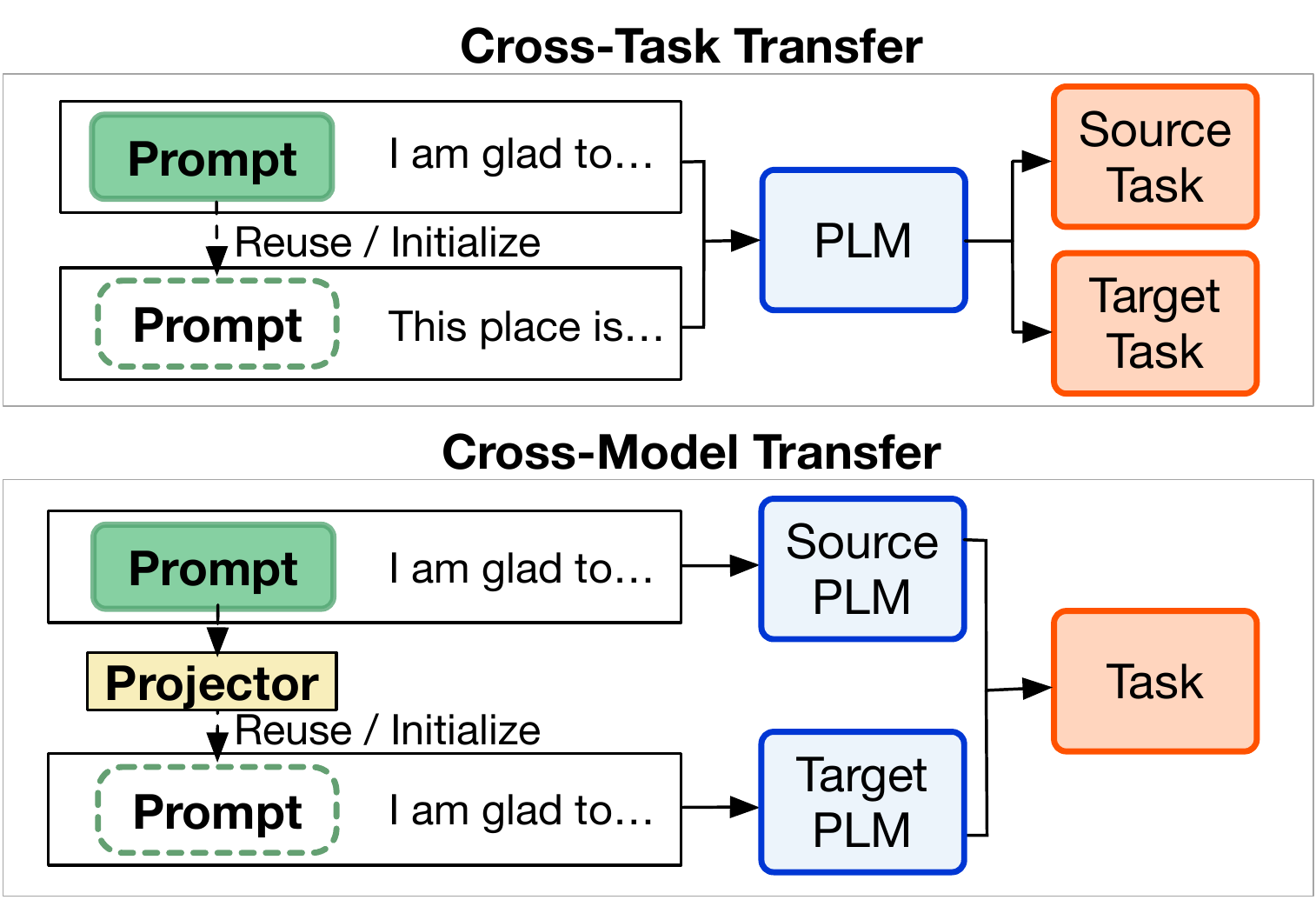}
\caption{We explore prompt transferring across different tasks (cross-task) and PLMs (cross-model) with directly reusing prompts and initializing prompt tuning.}
\label{fig:intro}
\end{figure}

Considering the extremely large scale of these state-of-the-art PLMs, conventional full-parameter fine-tuning methods become extremely expensive. Hence,  various parameter-efficient tuning methods \citep{houlsby2019parameter,ben2021bitfit,lester2021power,li-liang-2021-prefix,Liu2021GPTUT} are explored, among which prompt tuning (PT) has attracted broad research attention. PT prepends some \textit{soft prompts}, which are essentially learnable virtual tokens, into the input sequences and only trains them while keeping all the PLM's parameters fixed. The training objective is to generate desired outputs in the same way as the pre-training tasks. PT can match the downstream task performance of fine-tuning with only thousands of tunable parameters~\citep{lester2021power} when the PLM has billions of parameters.

\begin{figure}[!t]
\centering
\includegraphics[width=0.46\textwidth]{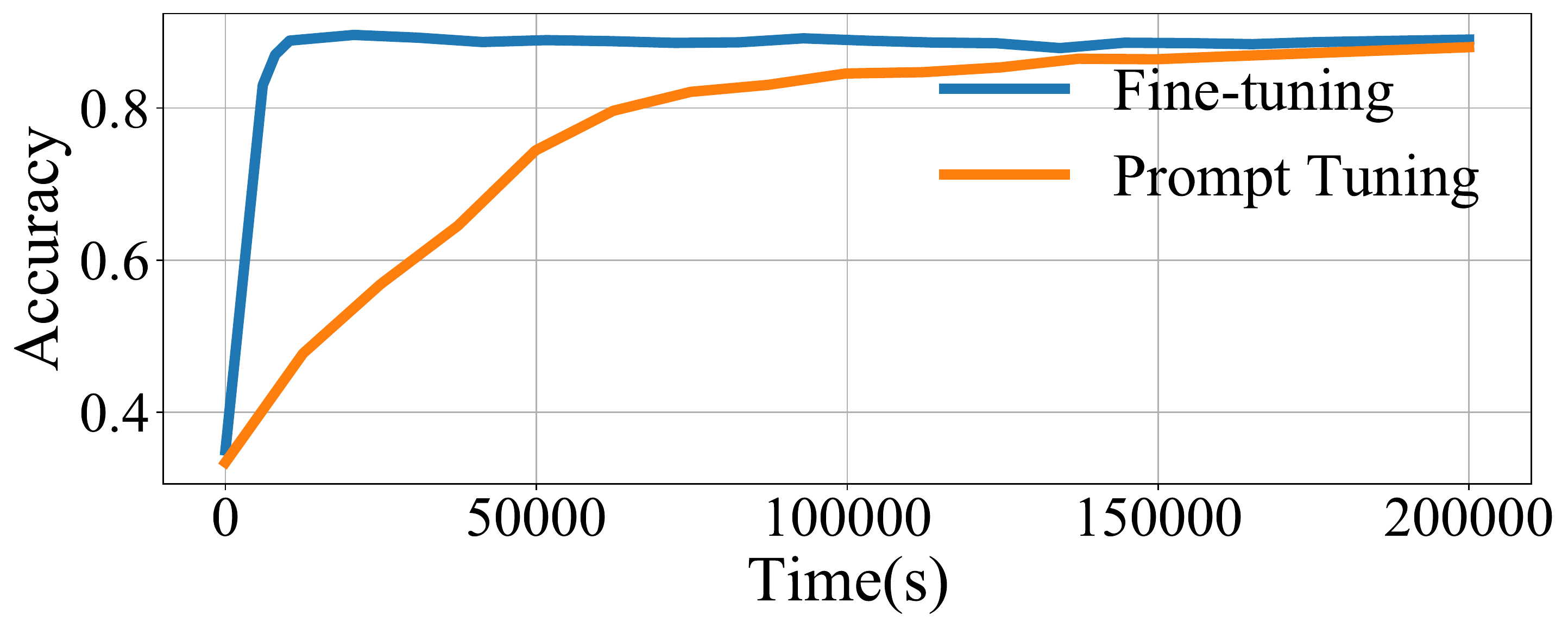}
\caption{Validation accuracies against training time of fine-tuning and PT for \RoBERTaLARGE on MNLI. PT takes much more training time.}
\label{fig:pt_slow_convergence}
\end{figure}

Although PT is an effective approach to utilizing extremely large PLMs, it requires much more training time than fine-tuning to reach the convergence as shown in Figure~\ref{fig:pt_slow_convergence}; hence, it is worthwhile to explore how to improve the efficiency of PT. In this work, we attempt to improve PT via \textbf{prompt transfer} across different tasks and models. Knowledge transfer across tasks~\citep{vu-etal-2020-exploring} and models~\citep{qin2021knowledge} have been widely used to improve the efficiency and effectiveness of NLP systems. Intuitively, soft prompts are the only tuned parameters in PT and thus shall concentrate the knowledge required to solve tasks conditioned on PLMs. Thus transferring the trained prompts is promising to accelerate PT.

\label{ssec:slow_convergence}

As shown in Figure~\ref{fig:intro}, we empirically analyze the transferability of prompts across different tasks (\textit{cross-task transfer} setting) and PLMs (\textit{cross-model transfer} setting) in this paper. The empirical analysis is conducted on $17$ NLP tasks of $6$ types and two representative PLM series: RoBERTa~\cite{liu2019roberta} and T5~\citep{raffel2020exploring}. In cross-task transfer, the prompt transfer can be done by directly reusing the trained prompts of the source task on the target task. However, in cross-model transfer, directly reusing prompts is intractable since the semantic spaces of different PLMs are inconsistent; hence, we develop various \textbf{prompt projectors} to project the soft prompts trained on the source PLM to the semantic space of the target PLM. We conduct two lines of experiments: (1) We investigate the \textbf{zero-shot transfer performance} and find that the transferability of prompts is influenced by task types. In cross-task transfer, the soft prompts can directly transfer to same-type tasks and achieve non-trivial performance, but poorly transfer to different-type tasks requiring different language skills. In cross-model transfer, we can successfully train a prompt projector with PT on a task, but the trained projector also only well generalizes to the same-type tasks of the projector-training task. (2) To accelerate PT, we propose to \textbf{transfer prompts with initialization}. In cross-task transfer, we start PT with the trained soft prompts of similar tasks as initialization. While in cross-model transfer, the initialization is the projected prompts of the same task trained on the source PLM. The two methods are dubbed as \TPTtask and \TPTmodel, which are short for transferable prompt tuning. Experiments show that they can both accelerate PT to some extent and also achieve a certain performance improvement.

Furthermore, we explore why can the prompts transfer and what decides their transferability. To this end, we design various prompt similarity metrics from different perspectives and examine how well they can serve as \textbf{transferability indicators}, i.e., how well they correlate with prompt transfer performance. Experiments find that our novel method of measuring prompt similarity via model activations in feed-forward layers is better correlated with prompt transferability than prompt embedding distance-based metrics. This suggests the prompts are essentially stimulating PLM's inner ability distributing among neurons to do specific NLP tasks, and future prompt transfer works should focus more on how the PLMs respond to different prompts' stimulation rather than the prompts' embedding properties.

To summarize, our contributions are three-fold: (1) We thoroughly analyze the transferability of prompts across different tasks and models, and show that improving PT with prompt transfer is possible and promising. (2) We propose to transfer prompts with initialization, which enhances both PT's efficiency and effectiveness. (3) We explore the effectiveness of various prompt similarity metrics serving as transferability indicators and demonstrate how the prompts stimulate PLMs to decide the transferability, which may facilitate further transferrable PT research.

%% file: sections/related_work.tex
\section{Related Work}
\label{label:related_work}

\paragraph{Prompt Tuning}
GPT-3~\citep{brown2020GPT3} demonstrates remarkable few-shot performance by prepending textual prompts before the inputs and thus helps the PLM to generate desired outputs of NLP tasks directly. Motivated by this, many works have tried to improve various NLP tasks by creating manually-crafted~\citep{schick-schutze-2021-exploiting,schick-schutze-2021-just,mishra2021reframing} or automatically-searched~\citep{jiang-etal-2020-know,shin-etal-2020-autoprompt,gao-etal-2021-making} \textit{hard prompts}, which are discrete tokens but not necessarily human-readable. Furthermore, \textit{soft prompts}~\citep{hambardzumyan2021warp,qin-eisner-2021-learning,zhong-etal-2021-factual,Liu2021GPTUT} are proposed, which are tuneable embeddings rather than tokens in the vocabularies and can be directly trained with task-specific supervision. \citet{lester2021power} demonstrate that prompt tuning (PT) method can match the performance of full-parameter fine-tuning when the PLM has billions of parameters. This suggests that PT is promising to utilize extremely large PLMs. However, the much more training time needed to reach the convergence makes PT inefficient. In this work, we show that prompt transfer can improve the effectiveness to some extent with knowledge transfer, and empirically analyze the transferability of prompts across tasks and PLMs.

\paragraph{Knowledge Transfer}
Cross-task knowledge transfer~\citep{ruder2017overview} has been a long-standing way to improve the effectiveness and efficiency of NLP systems. In the PLM era, some works propose to tune the PLMs on intermediate tasks~\citep{phang2019sentence,pruksachatkun-etal-2020-intermediate,gururangan2020don,wang-etal-2019-tell,vu-etal-2020-exploring,poth2021pretrain} before fine-tuning on specific target tasks to achieve certain benefits. \citet{vu-etal-2020-exploring} empirically analyze the transferability between tasks in this setting. 

These explorations are all for fine-tuning. Considering the potential of PT, we believe the transferability and knowledge transfer methods for PT are worth exploring. As a prior attempt, \citet{lester2021power} demonstrate that PT's cross-domain transferability is stronger than fine-tuning. 

Similar to our work, recent work~\citep{vu2021spot} also explores the cross-task transfer with prompt initialization and prompt similarity metrics based on cosine similarity. However, \citet{vu2021spot} focus on improving the effectiveness of PT but we attempt to improve the efficiency. Additionally, we explore more transferability indicators, especially the overlapping rate of activated neurons, and also investigate cross-model transfer, which is inspired by previous cross-model knowledge transfer works such as Net2Net~\citep{chen2016net2net}, knowledge distillation~\citep{44873} and knowledge inheritance~\citep{qin2021knowledge}.

%% file: sections/background.tex
\section{Preliminary}
\label{sec:preliminary}
Here we introduce the basic knowledge about PT (\cref{ssec:prompt_tuning}) as well as the downstream tasks (\cref{ssec:investigated_nlu_tasks}) and models (\cref{ssec:investigated_models}) investigated in experiments.

\subsection{Prompt Tuning}
\label{ssec:prompt_tuning}
In this work, we study the PT method that is capable of tuning large PLMs~\citep{lester2021power,Liu2021GPTUT}, i.e., we only explore the PT method freezing PLM parameters. PT prepends some virtual tokens, i.e., the \textit{soft prompts}, into the inputs of the PLM to provide knowledge about downstream tasks. The soft prompts are essentially tunable embedding vectors, which are trained with the objective enforcing the PLM to generate desired outputs of the downstream task in the same way of the pre-training objective.

Formally, given an input sequence with $n$ tokens $X = \{x_1, x_2,\ldots,x_{n}\}$, we first prepend $l$ randomly initialized soft prompts $P = \{\mathbf{p}_1,\mathbf{p}_2,\ldots,\mathbf{p}_l\}$ before them, where $\mathbf{p}_{i}\in \mathbb{R}^{d}$ is an embedding vector, and $d$ is the input dimension of the PLM. The training objective is to maximize the likelihood of decoding the desired output $y$:
\begin{equation}
\label{eq:continuous_prompt_tuning}
    \mathcal{L}=p(y|P,x_1,\ldots,x_n),
\end{equation}
where only $P$ is learnable. For the language understanding tasks, $y$ is the label token corresponding to the label of $X$. For the conditional generation tasks, $y$ is a sequence. Especially, for the models pre-trained with the masked language modeling objective like RoBERTa, we additionally prepend a special \texttt{[MASK]} token before the prompts and train the prompts to let the PLM fill $y$ into it.

\subsection{Investigated NLP Tasks}
\label{ssec:investigated_nlu_tasks}
To comprehensively study the prompt transferability across various NLP tasks, we involve $17$ diverse tasks, which can be divided into $6$ types: (1) \textbf{Sentiment Analysis (SA)}, including \texttt{IMDB} \cite{maas-EtAl:2011:ACL-HLT2011}, \texttt{SST-2} \cite{socher-etal-2013-recursive}, \texttt{laptop} \cite{pontiki-etal-2014-semeval}, \texttt{restaurant} \cite{pontiki-etal-2014-semeval}, Movie Rationales (\texttt{Movie}) \cite{zaidan-eisner-piatko-2008:nips} and TweetEval (\texttt{Tweet}) \cite{barbieri-etal-2020-Tweeteval}; (2) \textbf{Natural Language Inference~(NLI)}, including \texttt{MNLI} \citep{williams-etal-2018-broad}, \texttt{QNLI} \citep{wang2018glue} and \texttt{SNLI} \citep{bowman-etal-2015-large}; (3) \textbf{Ethical Judgment~(EJ)}, including \texttt{deontology} \cite{hendrycks2021aligning} and \texttt{justice} \cite{hendrycks2021aligning}; (4) \textbf{Paraphrase Identification~(PI)}, including \texttt{QQP} \citep{sharma2019natural} and \texttt{MRPC} \citep{dolan-brockett-2005-automatically}; (5) \textbf{Question Answering~(QA)}, including \texttt{SQuAD} \citep{rajpurkar-etal-2016-squad} and \texttt{NQ-Open} \citep{lee-etal-2019-latent}; (6) \textbf{Summarization~(SUM)}, including \texttt{Multi-News} \citep{fabbri-etal-2019-multi} and \texttt{SAMSum} \citep{gliwa-etal-2019-samsum}. Details for these tasks, evaluation metrics, label tokens, implementations are in \cref{sec:task_setup}.

\begin{figure*}[!t]
    \centering
    \subfigure[\RoBERTaLARGE]{
	    \includegraphics[width=0.482\textwidth]{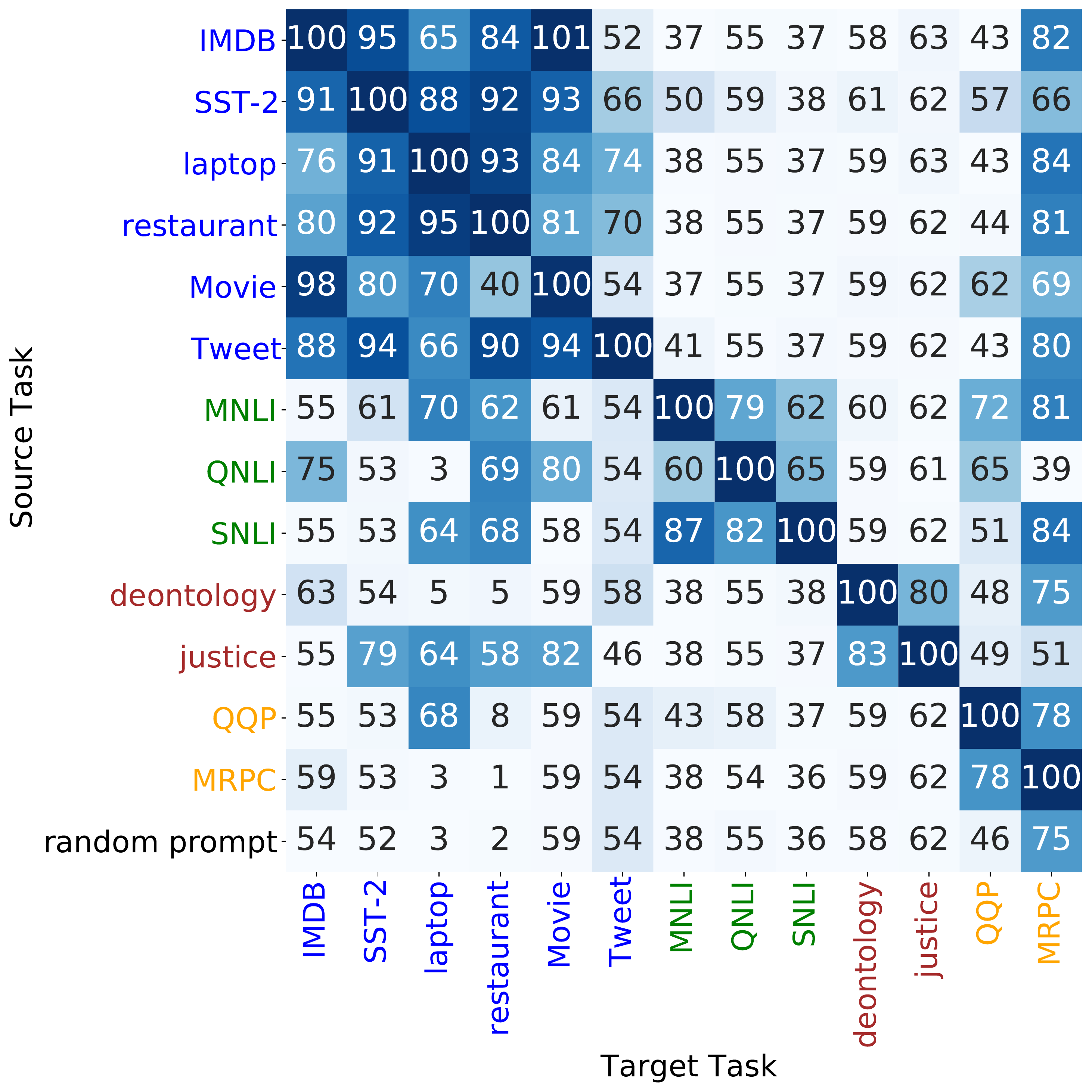}
    }
    \subfigure[\TXXL]{
	    \includegraphics[width=0.482\textwidth]{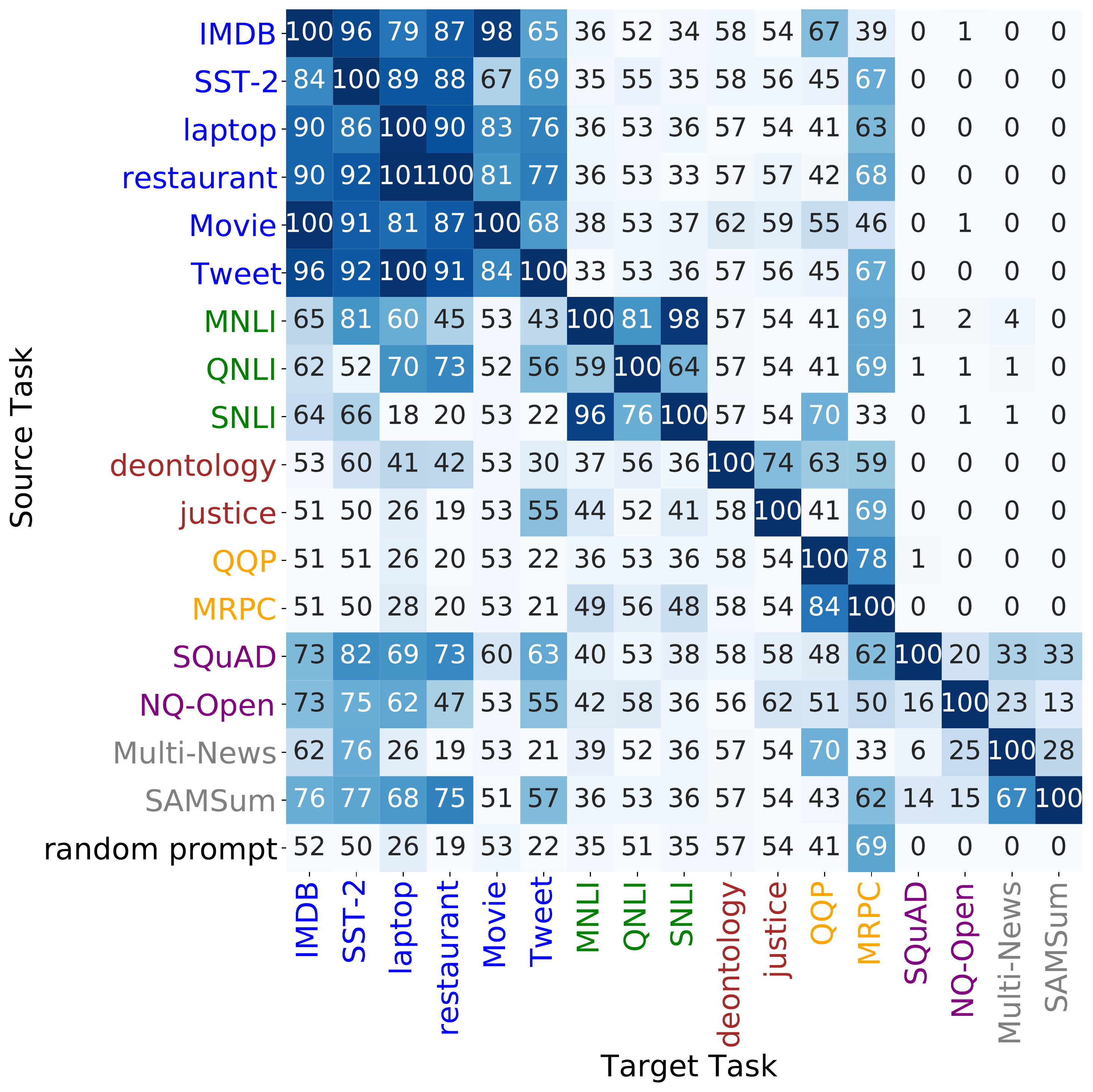}
    }
    \caption{Relative zero-shot transfer performance (zero-shot transfer performance / original PT performance) (\%) on the target tasks (columns) of the soft prompts trained on the source tasks (rows) for \RoBERTaLARGE and \TXXL. Colors of the task names indicate task types. \textcolor{blue}{Blue}: SA. \textcolor{OliveGreen}{Green}: NLI. \textcolor{Brown}{Brown}: EJ. \textcolor{YellowOrange}{Orange}: PI. \textcolor{Purple}{Purple}: QA. \textcolor{Gray}{Gray}: SUM. \textit{Random Prompt} of the last row means the soft prompts are randomly generated without any training.}
    \label{fig:zero-shot-task-transfer}
\end{figure*}

\subsection{Investigated Models}
\label{ssec:investigated_models}
We investigate prompt transferability for two series of PLMs: RoBERTa~\citep{liu2019roberta} and T5~\citep{raffel2020exploring}, which represent two mainstream pre-training types: masked language modeling and sequence-to-sequence pre-training. Considering RoBERTa can only predict a single token (or a fixed length of tokens) under prompt tuning paradigm, for the conditional generation tasks (QA and SUM) that output multiple tokens, we only investigate T5. We mainly report results for the two largest versions of PLMs, i.e., \RoBERTaLARGE and \TXXL. The more detailed results for the other sizes are attached in appendix.

%% file: sections/cross_task.tex
\section{Cross-Task Transfer}
\label{sec:cross-task}
We empirically study the cross-task transferability of soft prompts (\cref{ssec:cross_task_performance}) and try to improve the effectiveness and efficiency of PT with transfer (\cref{ssec:cross_task_initialization}). %All the experiments in this section are conducted on \RoBERTaBASE.

\begin{table*}[!t]
{\small
% \scalebox{0.59}
\begin{adjustbox}{max width=1\linewidth}
{
\setlength\tabcolsep{0.05em}
\begin{tabular}{l|*{6}{d{3.2}}|*{3}{d{3.2}}|*{2}{d{3.2}}|*{2}{d{3.2}}|*{2}{d{3.2}}|*{2}{d{3.2}}}
% \begin{tabular}{l|P{3.1}P{3.1}P{3.1}P{3.1}P{3.1}P{3.1}|P{3.1}P{3.1}P{3.1}|P{3.1}P{3.1}|P{3.1}P{3.1}|P{3.1}}
\toprule
  {\textbf{Task Type}} & \multicolumn{6}{c|}{\textbf{SA}} & \multicolumn{3}{c|}{\textbf{NLI}} & \multicolumn{2}{c|}{\textbf{EJ}} & \multicolumn{2}{c|}{\textbf{\makecell[c]{PI}}} & \multicolumn{2}{c|}{\textbf{\makecell[c]{QA}}} 
  & \multicolumn{2}{c}{\textbf{\makecell[c]{SUM}}}
  %Summarization
  %\multicolumn{1}{c}{\textbf{Average}}
  \\
  \midrule
  {\textbf{Task}} & \mc{IMDB} & \mc{SST-2} & \mc{laptop} & \mc{restaurant} & \mc{Movie} & \mcl{Tweet} & \mc{MNLI} & \mc{QNLI} & \mcl{SNLI} & \mc{deontology} & \mcl{justice} & \mc{QQP} & \mcl{MRPC} & \mc{SQuAD} & \mcl{NQ-Open} & \mc{Multi-News} & \mc{SAMSum}\\ 
  \midrule
{\textbf{Metric}} & \mc{Acc.} & \mc{Acc.} & \mc{Acc.} & \mc{Acc.} & \mc{Acc.} & \mcl{Acc.} & \mc{Acc.} & \mc{Acc.} & \mcl{Acc.} & \mc{Acc.} & \mcl{Acc.} & \mc{Acc.} & \mcl{Acc.} & \mc{F1} & \mcl{F1} & \mc{ROUGE-L} & \mc{ROUGE-L}\\ 
\midrule
\multicolumn{18}{c}{\RoBERTaLARGE} \\
\midrule
{\textbf{Performance (PT) (\%)}} & 92.2 & 96.1 & 76.4 & 83.7 & 84.9 & \mcl{{\textbf{76.1}}} & 87.3 & 92.4 & \mcl{\textbf{91.9}} & \mc{\textbf{85.6}} & \mcl{\textbf{81.0}} & \mc{\textbf{88.9}} & \mcl{\textbf{81.2}} & \mc{N/A} & \mcl{N/A} & \mc{N/A} & \mc{N/A} \\ 
{\textbf{Performance (\TPTtask) (\%)}} & \mc{\textbf{92.4}} & \mc{\textbf{96.3}} & \mc{\textbf{79.1}} & \mc{\textbf{85.8}} & \mc{\textbf{85.1}} & \mcl{\textbf{76.1}} & \mc{\textbf{87.9}} & \mc{\textbf{93.1}} & \mcl{\textbf{91.9}} & \mc{\textbf{85.6}} & 78.2 & 86.1 & 79.2 & \mc{N/A} & \mcl{N/A} & \mc{N/A} &  \mc{N/A}\\
\midrule
{\textbf{Convergence Speedup}} & 1.7 & 1.1 & 1.0 & 1.9 & 1.2 & 0.9 & 1.2 & 1.2 & 1.3 & 0.9 & 0.7 & 0.8 & 0.9 & \mc{N/A} & \mcl{N/A} & \mc{N/A} &  \mc{N/A}\\
{\textbf{Comparable-result Speedup}} & 2.5 & 2.4 & 1.0 & 3.8 & 1.5 & 1.3 & 1.1 & 2.3 & 1.0 & 0.9 & \mcl{N/A} & \mc{N/A} & \mcl{N/A} & \mc{N/A} & \mcl{N/A} & \mc{N/A} &  \mc{N/A}\\
\midrule
\multicolumn{18}{c}{\TXXL} \\
\midrule
{\textbf{Performance ($\mathbf{PT}$) (\%)}} & 96.5 & 97.4 & 76.6 & \mc{\textbf{90.1}} & \mc{\textbf{97.9}} & 76.2 & 90.5 & 95.2 & 93.4 & 87.0 & \mcl{\textbf{92.5}} & 90.0 & 86.3 & \mc{\textbf{86.3}} & 20.8 &  29.2 &  45.8\\
{\textbf{Performance (\TPTtask) (\%)}} & \mc{\textbf{96.6}} & \mc{\textbf{97.8}} & \mc{\textbf{84.2}} & 88.6 & 97.5 & \mcl{\textbf{77.0}} & \mc{\textbf{92.0}} & \mc{\textbf{96.2}} & \mcl{\textbf{94.0}} & \mc{\textbf{95.3}} & 90.7 & \mc{\textbf{90.9}} & \mcl{\textbf{89.0}} & 85.9 & \mcl{\textbf{21.3}} & \mc{\textbf{29.3}} &  \mc{\textbf{46.8}}\\
\midrule
{\textbf{Convergence Speedup}} & 1.2 & 49.7 & 2.2 & 1.1 & 3.9 & 1.4 & 12.5 & 24.9 & 49.9 & 29.8 &1.5 & 1.0 & 3.3 & 1.1 & 1.0 & 2.0 &  2.0 \\
{\textbf{Comparable-result Speedup}} & 1.2 & 48.9 & 219.8 & \mc{N/A} & \mc{N/A} & 1.5 & 12.5 & 29.9 & 49.9 & 29.9 & \mcl{N/A} & 1.0 & 5.0 & \mc{N/A} & 1.0 &  2.0 &  2.5\\
\bottomrule
\end{tabular}
}
\end{adjustbox}
\caption{Performance on $17$ NLP tasks of vanilla prompt tuning ($\mathbf{PT}$) and prompt tuning with transferring initialization (\TPTtask), which initialize PT with the one performing best in zero-shot transfer, as well as the convergence speedup (the quotient of the training steps of $\mathbf{PT}$ by the training time of \TPTtask reaching convergence) and comparable-result speedup (the quotient of the training time of $\mathbf{PT}$ by the training time of \TPTtask achieving comparable performance to $\mathbf{PT}$). N/A represents the tasks that \RoBERTaLARGE cannot conduct, or we fail to speed up training with \TPTtask.}
\label{table:initalize}
}
\end{table*} %Xiaozhi: add roberta large and bert base results here, change into Acc_std format

\subsection{Zero-shot Transfer Performance}
\label{ssec:cross_task_performance}
To study the cross-task transferability, we first examine PT's zero-shot transfer performance, i.e., we conduct PT on a source task, then directly reuse the trained prompts on other target tasks and evaluate their performance. 
The results are shown in Figure~\ref{fig:zero-shot-task-transfer}\footnote{More results on other PLMs are left in \cref{ssec:appendix_relative_performance_of_zero-shot_transferability_on_various_plms}.}, from which we can observe that: (1) For the tasks within the same type, transferring soft prompts between them can generally perform well and may even outperform vanilla PT on the target task, especially when the source task has more data (the case of transferring from \texttt{IMDB} to \texttt{Movie} in Figure~\ref{fig:zero-shot-task-transfer}~(a) and transferring from \texttt{restaurant} to \texttt{laptop} in Figure~\ref{fig:zero-shot-task-transfer}~(b)), which demonstrates that it is promising to improve PT's effectiveness and efficiency with knowledge transfer from similar tasks. (2) For the tasks of different types, the transferability of soft prompts among them is generally poor, and transferring soft prompts often achieve similar performance to randomly initialized prompts. 
 
(3) However, some tasks can transfer to different-type tasks to some extent, such as the QA and SUM tasks to SA tasks in Figure~\ref{fig:zero-shot-task-transfer}~(b). To understand this, it is worthwhile to explore what controls the transferability between prompts, and we do some preliminary study in \cref{ssec:cross-task_prompt_similarity}.

\subsection{Transfer with Initialization}
\label{ssec:cross_task_initialization}
To improve the effectiveness and efficiency of PT with cross-task transfer, we explore a cross-task transferable prompt tuning (\TPTtask) method, which initializes soft prompts with well-trained prompts of the most similar task and then starts PT.%We propose to initialize the soft prompts with well-trained soft prompts of similar tasks before starting prompt tuning and observe whether these initalizations can accelerate training and improve final performances.

For a target task, we start \TPTtask with trained prompts of the source task achieving the best zero-shot transfer performance in Figure~\ref{fig:zero-shot-task-transfer}. From the results of the performance and training time comparisons\footnote{Training time comparisons are left in \cref{ssec:appendix_details_of_tpttask_speedup}.} in Table~\ref{table:initalize}, we can see \TPTtask can mostly achieve better or comparable performance to vanilla PT starting from random initialization, and \TPTtask generally takes less training time.

%% file: sections/cross_model.tex
\section{Cross-Model Transfer}
\label{sec:cross-model}
We further study the cross-model transferability of soft prompts. Intuitively, cross-model transfer allows us to train prompts on a small and computationally efficient PLM and use them on a massive and computationally expensive PLM, which will be much more efficient and environment-friendly. 
We investigate the feasibility of cross-model transfer on transferring from a source PLM (\RoBERTaLARGE) to a larger and heterogeneous target PLM (\TXXL), which shall be the most difficult setting. Appendix~\ref{ssec:appendix_cross-model_transfer} shows the experimental results of other settings. Directly reusing trained soft prompts between different PLMs is infeasible since their embedding spaces are different. Hence, we investigate how to do cross-model prompt projection (\cref{ssec:cross-model_projector}) and see the transfer performance (\cref{ssec:zero-shot_transferability_cross-model}). Furthermore, we explore to improve PT with cross-model transfer initialization (\cref{ssec:transferring_with_initialization_cross-model}).

\begin{table*}[!t]
\begin{center} 
%\small
\begin{adjustbox}{max width=1\linewidth}
{
\setlength\tabcolsep{0.2em}

\begin{tabular}{c|cl|*{6}{d{3.2}}|*{3}{d{3.2}}|*{2}{d{3.2}}|*{2}{d{3.2}}}
\toprule
\multicolumn{3}{c|}{\multirow{2}{*}{\textbf{Method}}} &
\multicolumn{6}{c|}{\textbf{SA}} & \multicolumn{3}{c|}{\textbf{NLI}} & \multicolumn{2}{c|}{\textbf{EJ}} & \multicolumn{2}{c}{\textbf{PI}} \\ \cline{4-16} 
\multicolumn{3}{c|}{} & \mc{IMDB} & \mc{SST-2} & \mc{laptop} & \mc{restaurant} & \mc{Movie} & \mcl{Tweet} & \mc{MNLI} & \mc{QNLI} & \mcl{SNLI} & \mc{deontology} & \mcl{justice} & \mc{QQP} & \mc{MRPC}\\ 
\midrule
\multicolumn{3}{l|}{PT on \TXXL} & 96.5 & 97.4 & 76.6 & 88.1 & 97.9 & 72.5 & 90.5 & 95.2 & 93.4 & 87.0 & 92.5 & 90.0 & 86.3\\
\midrule
\multicolumn{3}{l|}{Random Prompt} & 49.7 & 49.0 & 19.8 & 17.0 & 51.6 & 15.5 & 31.8 & 49.3 & 31.9 & 51.3 & 50.0 & 36.4 & 67.0 \\
%\midrule
\midrule
\multicolumn{16}{c}{(a) Zero-shot Transfer Performance (\%)} \\
\midrule
\multicolumn{1}{l|}{\multirow{2}{*}{\texttt{laptop}}} & \multicolumn{2}{l|}{Distance Minimizing}  & 49.6 & 49.0 & 76.6 & 17.5 & 51.5 & 14.4 & 31.8 & 48.1 & 32.8 & 53.3 & 49.9 & 36.8 & 66.6 \\
{} & \multicolumn{2}{l|}{Task Tuning} & \mc{\textbf{82.9}} & \mc{\textbf{89.3}} & \mc{\textbf{80.3}} & \mc{\textbf{85.7}} & \mc{\textbf{78.6}} & \mcl{\textbf{58.4}} & 32.4 & 50.7 & 33.6 & 54.9 & 51.6 & 33.9 & 63.7 \\
\midrule
\multicolumn{1}{l|}{\multirow{2}{*}{\texttt{MNLI}}} &
\multicolumn{2}{l|}{Distance Minimizing}  & 49.6 & 50.1 & 19.8 & 18.3 & 51.2 & 15.0 & \mc{\textbf{90.5}} & 49.0 & 32.9 & 50.3 & 49.0 & 36.8 & 65.6 \\
{} & \multicolumn{2}{l|}{Task Tuning} & 49.7 & 48.8 & 19.8 & 17.0 & 51.6 & 16.0 & 89.8 & \mc{\textbf{82.7}} & \mcl{\textbf{88.2}} & 49.7 & 50.0 & 36.8 & 67.7 \\ 
%\midrule
\midrule
\multicolumn{16}{c}{(b) Transfer with Initialization (\TPTmodel)} \\
\midrule
\multicolumn{1}{l|}{\multirow{3}{*}{\texttt{laptop}}} & \multicolumn{2}{l|}{Performance (\%)}  & 96.5 & 97.4 & 82.9 & 90.3 & 97.4 & 74.4 & 91.0 & 95.4 & 93.4 & 92.5 & 92.5 & 90.0 & 87.9 \\
{} & \multicolumn{2}{l|}{Convergence Speedup}  & 1.1 & 1.7 & 1.9 & 1.3 & 0.6 & 1.3	& 0.9 & 0.9 & 1.0 & 1.0 & 0.7 & 1.1 & 1.1 \\
{} & \multicolumn{2}{l|}{Comparable-result Speedup} & 1.0 & 19.0 & 16.0 & 6.0 & \mc{N/A} & 2.2 & 3.6 & 1.1 & 6.0 & 6.0 & 0.9 & 1.8 & 3.4\\

\midrule
\multicolumn{1}{l|}{\multirow{3}{*}{\texttt{MNLI}}} &  \multicolumn{2}{l|}{Performance (\%)} & 96.5 & 97.4 & 82.7 & 88.5 & 95.8 & 74.7 & 91.2 & 95.9 & 93.5 & 94.6 & 92.5 & 90.0 & 87.7 \\
{} &
\multicolumn{2}{l|}{Convergence Speedup}  &  1.0 & 1.6 & 1.8 & 0.9 & 0.4 & 1.3 & 1.0 & 1.1 & 1.4 & 2.0 & 1.7 & 0.9 & 0.9 \\
{} & \multicolumn{2}{l|}{Comparable-result Speedup} &  1.0 & 18.0 & 15.0 & 1.6 & \mc{N/A} & 1.5 & 18.0 & 20.0 & 30.0 & 7.5 & 5.0 & 1.5 &1.9 \\
\bottomrule
\end{tabular}
}
\end{adjustbox}
\caption{Cross-model prompt transfer (\RoBERTaLARGE to \TXXL) results, including non-transfer baselines (vanilla PT and randomly generated prompts), zero-shot transfer performance of various projectors, and \TPTmodel results (performance, convergence speedup, and comparable-result speedup similar to Table~\ref{table:initalize}). \TPTmodel adopts the Task Tuning projectors to project the soft prompts.}
\label{table:robertalarge_to_t5xxl}
\end{center} 
\end{table*}

\subsection{Cross-Model Prompt Projection}
\label{ssec:cross-model_projector}
To project the trained soft prompts of a PLM to the semantic space of a different PLM, we train projectors with various objectives and examine their effectiveness. A good way to train the cross-model projectors may need some task-specific supervisions, but the trained projector shall generalize to different tasks so that the efficiency for learning the new tasks on the target model could be improved.

Formally, given the prompt of the source PLM $P^s=\{\mathbf{p}_{1}^{s},\ldots,\mathbf{p}_{l}^{s}\}$, we concatenate the $l$ virtual tokens into a unified vector $\mathbf{P}^{s}\in\mathbb{R}^{l d_s}$. The projector $\mathbf{Proj(\cdot)}$ is to project it to $\mathbf{\tilde{P}}^{s} \in\mathbb{R}^{ld_t}$ in the semantic space of the target PLM, where $d_s$ and $d_t$ are the input embedding dimensions of the source and target PLM, respectively. We parameterize the projector with a two-layer perceptron as follows: 
\begin{equation}\label{eq:transfer_projector}
    \mathbf{\tilde{P}}^{s}\! =\! \mathbf{Proj}(\mathbf{P}^{s})\! =\! \mathbf{W}_{2}(\sigma(\mathbf{P}^{s}\mathbf{W}_{1}\!+\!\mathbf{b}_{1}))\!+\!\mathbf{b}_{2},
\end{equation}
where $\mathbf{W}_{1}\in\mathbb{R}^{d_h\times l d_s},\mathbf{W}_{2}\in \mathbb{R}^{l d_t \times d_h}$ are trainable matrices, $\mathbf{b}_1\in \mathbb{R}^{d_h},\mathbf{b}_2 \in \mathbb{R}^{l d_t}$ are biases, $\sigma$ is a non-linear activation function. We investigate two learning objectives to train the projector\footnote{More projector-training details are left in \cref{ssec:appendix_implementation_details_of_projectors}.}:

\paragraph{Distance Minimizing} We firstly try to learn cross-model projections by minimizing the distance between the projected prompt and the parallel prompt $\mathbf{P}^{t}$ originally trained on the target PLM with the same task, i.e., the training objective is to minimize their $L_{2}$-distance $\Vert \mathbf{Proj}(\mathbf{P}^{s}) - \mathbf{P}^{t}  \Vert_{2}$.

\paragraph{Task Tuning} We then try to train the cross-model projector with task-specific supervision signals on the target PLM. Specifically, we directly tune the projected prompts on some tasks and back propagate the supervision signals to train the projector weights, so that the projector can learn how to stimulate the target PLM and thus may generalize to transfer the prompts of other tasks.

These methods rely on some tasks (parallel trained soft prompts or training data) to train the projector. The projector learning methods are agnostic to the specific training tasks used, and we choose \texttt{laptop} and \texttt{MNLI} in experiments.

\subsection{Zero-shot Transfer Performance}
\label{ssec:zero-shot_transferability_cross-model}

The zero-shot transfer performance of various projector-learning methods are shown in Table \ref{table:robertalarge_to_t5xxl}\footnote{More results on other PLMs are left in appendix \ref{ssec:appendix_zero-shot_transfer_performance_on_various_plms}.} (a). We can observe that: (1) \textbf{Distance Minimizing} works well to transfer the prompts of the projector-training task, but falls back to random performance on the other unseen tasks, which is not practically usable. This is consistent with our findings in~\cref{ssec:cross-task_prompt_similarity} that the embedding distances do not strongly correlate to prompt transferability. (2) \textbf{Task Tuning} performs better and successfully generalizes to same-type unseen tasks of the projector-training tasks (e.g. NLI tasks for the projectors trained with \texttt{MNLI}), which proves the feasibility of practical cross-model prompt transfer. (3) The projectors trained with \textbf{Task Tuning} still cannot work for different-type tasks, which may be limited by the cross-task prompt transferability investigated in \cref{ssec:cross_task_performance}. This urges further attention to developing universal cross-model projections.

\subsection{Transfer with Initialization}
\label{ssec:transferring_with_initialization_cross-model}

Similar to~\cref{ssec:cross_task_initialization}, we further study whether the projected soft prompts can initialize PT on the target PLM and accelerate training as well as improve performance. We propose cross-model transferable prompt tuning, \TPTmodel, which adopts the \textbf{Task Tuning} projectors to project the soft prompts trained on the source PLM into the target PLM and initialize PT with the projected prompts.

The performance and speedup are shown in Table \ref{table:robertalarge_to_t5xxl} (b). We can see that, for the tasks within the same type of the projector-training task, compared to vanilla PT, \TPTmodel can mostly achieve comparable or better performance with much less training time, which demonstrates that practical cross-model prompt transfer is promising for improving the efficiency and effectiveness of PT.

%% file: sections/similarity.tex
\section{Exploring Transferability Indicator}
\label{ssec:cross-task_prompt_similarity}
Based on the positive results in cross-task and cross-model transfer, we explore why the soft prompts can transfer across tasks and what decides the transferability between them, which may shed light on the mechanisms behind PT and help to design transferable PT methods. We explore various \textbf{prompt similarity metrics} and examine how well do they align with the zero-shot transfer performance. If a similarity metric can well indicate transferability, it suggests the factors considered in designing this metric decide the prompt transferability. Moreover, the prompt similarity metrics can qualify task similarities using the trained soft prompts as task embeddings and may help in developing cross-task transfer methods. As a straightforward example, if we build a \textit{prompt warehouse} containing prompts of diverse tasks, we can retrieve prompts of similar tasks for a new task with a certain similarity metric and better improve PT with \TPTtask. 

\subsection{Prompt Similarity Metric}
We explore the following two kinds of metrics:

\paragraph{Embedding Similarity}
We firstly regard the trained soft prompts as only embeddings in the vector space and calculate their \textit{Euclidean similarity} and \textit{cosine similarity}, among which cosine similarity is also explored by \citet{vu2021spot}.

Given two groups of trained prompts containing $l$ virtual tokens: $P^{t_1}=\{\mathbf{p}^{t_1}_{1},\ldots,\mathbf{p}^{t_1}_{l}\}$ and $P^{t_2}=\{\mathbf{p}^{t_2}_{1},\ldots,\mathbf{p}^{t_2}_{l}\}$, which correspond to tasks $t_1$ and $t_2$. Firstly, we concatenate the $l$ virtual tokens for each group and get two concatenation embeddings $\mathbf{P}^{t_1},\mathbf{P}^{t_2}\in\mathbb{R}^{ld}$, then we compute Euclidean similarity and cosine similarity of them:
\begin{equation}
    \small
    \begin{aligned}
        \mathrm{E}_{\mathrm{concat}}(P^{t_1},P^{t_2}) &=\frac{1}{1+\Vert\mathbf{P}^{t_1}-\mathbf{P}^{t_2}\Vert}, \\
        \mathrm{C}_{\mathrm{concat}}(P^{t_1},P^{t_2}) &=\frac{\mathbf{P}^{t_1}\cdot\mathbf{P}^{t_2}}{\Vert\mathbf{P}^{t_1}\Vert\Vert\mathbf{P}^{t_2}\Vert}.
    \end{aligned}
\end{equation}

We further explore a simple way to make the metrics invariant to token positions. We compute Euclidean distances and cosine similarities for every virtual token pairs in the two groups and use the averaged results in the final similarity metrics:
\begin{equation}
    \small
    \begin{aligned}
        \mathrm{E}_{\mathrm{average}}(P^{t_1},P^{t_2}) &=\frac{1}{1+\frac{\displaystyle 1}{\displaystyle l^2}\sum\limits_{i=1}^{l}\sum\limits_{j=1}^{l}\Vert\mathbf{p}^{t_1}_{i}-\mathbf{p}^{t_2}_{j}\Vert}, \\
        \mathrm{C}_{\mathrm{average}}(P^{t_1},P^{t_2}) &=\frac{\displaystyle 1}{\displaystyle l^2}\sum_{i=1}^{l}\sum_{j=1}^{l}\frac{\mathbf{p}^{t_1}_{i}\cdot\mathbf{p}^{t_2}_{j}}{\Vert\mathbf{p}^{t_1}_{i}\Vert\Vert\mathbf{p}^{t_2}_{j}\Vert}.
    \end{aligned}
\end{equation}

\paragraph{Model Stimulation Similarity}
In the second way, we depict their similarities based on how they \textit{stimulate the PLMs}, i.e., we examine the similarities between the responses of PLMs to the two soft prompts. Motivated by \citet{geva2021transformer} and \citet{dai2021knowledge}, which both find that the activation of the neurons in the feed-forward layers of Transformers~\cite{NIPS2017_3f5ee243} corresponds to specific model behaviors, we propose to use the \textit{overlapping rate of activated neurons} as a similarity metric of prompts. Specifically, the feed-forward network $\mathrm{FFN}(\cdot)$ in a Transformer layer is: 
\begin{equation}
\begin{aligned}
%\mathrm{FFN}(\mathbf{x}) = f(\mathbf{x} W_1^{\top}+b_1) W_2 + b_2,
\mathrm{FFN}(\mathbf{x}) = \max(\mathbf{x} \mathbf{W}_{1}^{\top}+\mathbf{b}_1, \mathbf{0}) \mathbf{W}_{2} + \mathbf{b}_2,
\end{aligned}
\label{eq:neuron}
\end{equation}
where $\mathbf{x}\in\mathbb{R}^{d}$ is the input embedding, %$f(\cdot)$ is a non-linear activation function, 
$\mathbf{W}_{1},\mathbf{W}_{2}\in\mathbb{R}^{d_{m} \times d}$ are trainable matrices, and $\mathbf{b}_1, \mathbf{b}_2$ are bias vectors. The $\max(\mathbf{x} \mathbf{W}_{1}^\top+\mathbf{b}_1,\mathbf{0})$ can be regarded as the non-negative activation values for $d_{m}$ hidden neurons \cite{geva2021transformer}. We then change all the positive elements of $\max(\mathbf{x} \mathbf{W}_1^\top+\mathbf{b}_1,\mathbf{0})$ to $1$ and get the one-hot activation state vector $\mathbf{s}$.

We feed an input sequence $\{P,$\texttt{<s>}$\}$ into the PLMs, where \texttt{<s>} is the special token indicating the start of a sentence. For RoBERTa, a \texttt{[MASK]} is additional prepended. This sequence is in the format of PT inputs but without specific input sentences. 

We use the activation states of the positions used to decode outputs, which shall be more task-specific. Specifically, for T5, we use the decoder module's activation states at the first position. For RoBERTa, we use the activation states of \texttt{[MASK]}. Finally, we concatenate the activation states of PLM's $L$ layers to get the overall activation states:

\begin{equation}
\begin{aligned}
\mathrm{AS}({P}) =
[\mathbf{s}_{1};\mathbf{s}_{2};...;\mathbf{s}_{L}].
\end{aligned}
\label{eq:ap}
\end{equation}

\begin{table}[!t]
\begin{center} 
 \small
%\scalebox{0.9}
{
{

\begin{tabular}{l|l|*{2}{d{2.2}}}
\toprule
\multicolumn{1}{c|}{\textbf{Model}} & \multicolumn{1}{c|}{\textbf{Metric}} & \multicolumn{1}{c}{\textbf{\begin{tabular}[c]{@{}c@{}}Same\\ Task\end{tabular}}} & \multicolumn{1}{c}{\textbf{\begin{tabular}[c]{@{}c@{}}Different\\ Tasks\end{tabular}}} \\ \midrule
\multirow{5}{*}{\RoBERTaLARGE} & $\mathrm{E}_{\mathrm{concat}}$ & 9.4 & 6.8 \\
& $\mathrm{E}_{\mathrm{average}}$ & 41.6 & 37.6 \\
& $\mathrm{C}_{\mathrm{concat}}$ & 47.6 & 31.7\\
& $\mathrm{C}_{\mathrm{average}}$ & 1.7 & 1.1\\
& ON & 39.4 &  21.4\\ \midrule
\multirow{5}{*}{\TXXL} & $\mathrm{E}_{\mathrm{concat}}$ & 0.5 & 0.3 \\
& $\mathrm{E}_{\mathrm{average}}$ & 4.0 & 3.4\\
& $\mathrm{C}_{\mathrm{concat}}$ & 29.4 & 3.4 \\
& $\mathrm{C}_{\mathrm{average}}$ & 4.0 & 2.1\\
& ON & 62.0 & 46.1\\ 
\bottomrule
\end{tabular}
}}
\caption{The average values (\%) of the $5$ similarity metrics for prompt pairs of the same task (trained with $3$ different random seeds) and different tasks.}
\label{table:similiarty_metrics}
\end{center} 
\end{table}

We can only retrieve the activation states of a part of layers in the similarity computation. In experiments, we find that the higher layers tend to be more task-specific, which is consistent with the probing results~\citep{Liu2019LinguisticKA}. Hence we use the activation states of the top $3$ layers\footnote{More results about the different layers's performance are left in~\cref{ssec:appendix_overlapping_percentages_of_activated_neurons}.} in experiments below.
%and more results about the different layers's performance are left in~\cref{ssec:appendix_overlapping_percentages_of_activated_neurons}. 
We calculate the overlapping rate of activated neurons $\mathrm{ON}({P}^{t_1},{P}^{t_2})$ between the trained soft prompts of task $t_1$ and $t_2$ with the cosine similarity: 
\begin{equation}
\begin{aligned}
    \mathrm{ON}({P}^{t_1},{P}^{t_2}) = \frac{\mathrm{AS}(P^{t_1})\cdot\mathrm{AS}(P^{t_2})}{\Vert\mathrm{AS}(P^{t_1})\Vert\Vert\mathrm{AS}(P^{t_2})\Vert}.
    %cos(\mathrm{AP}(\mathbf{P}_{i}),\mathrm{AP}(\mathbf{P}_{j})).
\end{aligned}
\label{eq:overlap_ap}
\end{equation}

\begin{table}[!t]
\begin{center} 
\small
%\scalebox{0.9}
{
{
\begin{tabular}{l|*{2}{d{3.2}} }
\toprule
% \multirow{2}{*}{\textbf{Metric}} & \multicolumn{2}{c}{\textbf{Model}} \\ 
% \cline{2-3} 
\multirow{1}{*}{\textbf{Metric}} & \multicolumn{1}{c}{\RoBERTaLARGE} &  \multicolumn{1}{c}{\TXXL} \\ 
% {} & \multicolumn{1}{c}{\RoBERTaLARGE} & \multicolumn{1}{c}{\TXXL} \\ 
\midrule
$\mathrm{E}_{\mathrm{concat}}$  & 22.6 & 12.9 \\
$\mathrm{E}_{\mathrm{average}}$ & 2.8 & -2.5 \\
$\mathrm{C}_{\mathrm{concat}}$ & 24.8 & 31.6 \\
$\mathrm{C}_{\mathrm{average}}$ & 44.7 & 33.5 \\
$\mathrm{ON}$ & \mc{\textbf{49.7} } & \mc{\textbf{36.9}}\\  \bottomrule
\end{tabular}
}}
\caption{The Spearman's rank correlation scores (\%) between various similarity metrics and cross-task zero-shot transfer performance of soft prompts.} 
\label{table:prompt_similiarty_measure}
\end{center} 
\end{table}
%1 indicates the bottom layer and 12 indicates the top layer.

\begin{figure}[!t]
\centering
\includegraphics[width=0.47\textwidth]{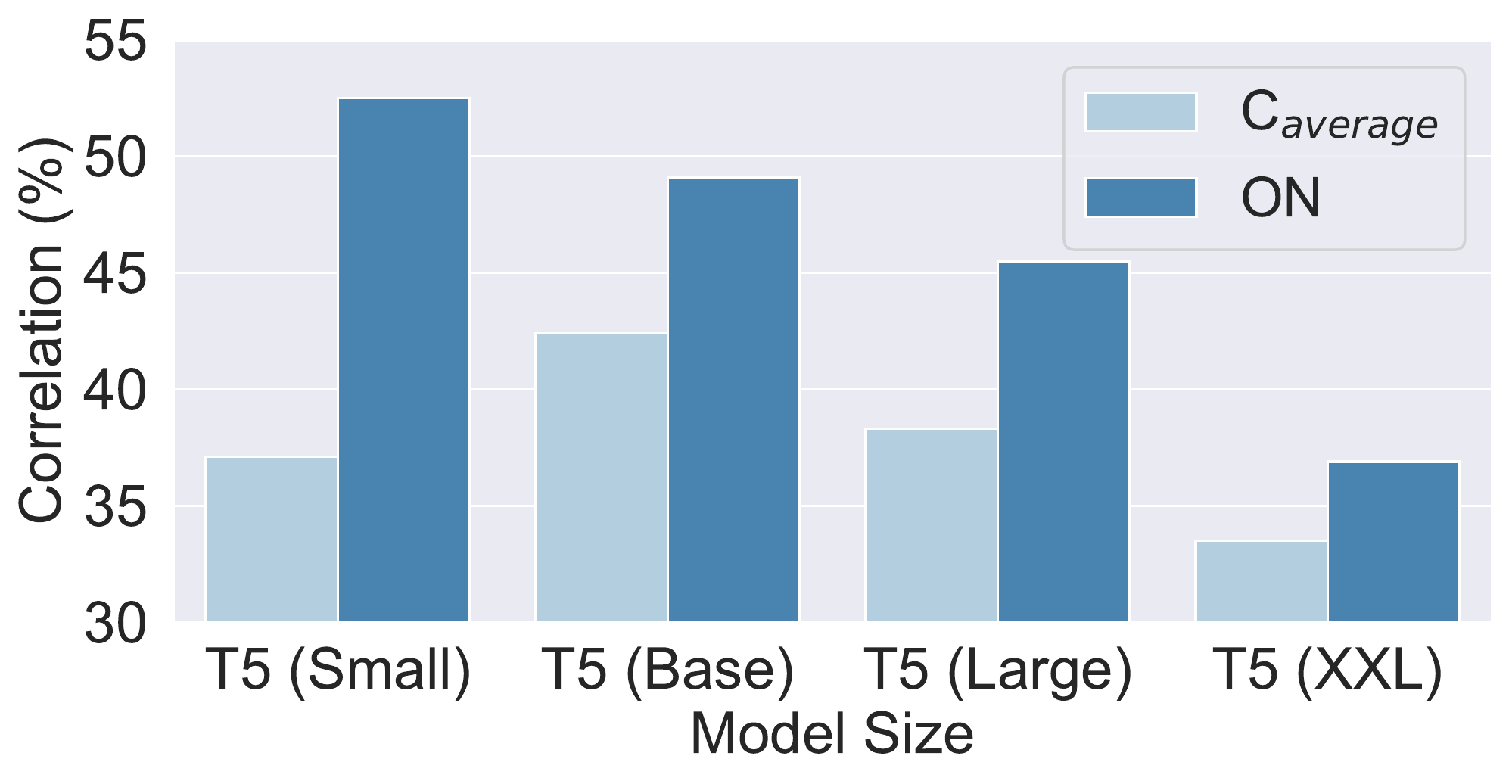}
\caption{Spearman's correlation scores of $\mathrm{ON}$ and $\mathrm{C}_{\mathrm{average}}$ with cross-task zero-shot transfer performance change along with the parameter size of T5.}
\label{fig:model_size_colrelation}
\end{figure}

\subsection{Experimental Results}
% \paragraph{Overlapping percentages of activated neurons can better present the task/prompt similarity.} 
To evaluate the effectiveness of the above similarity metrics of soft prompts, we (\romannumeral1) test whether the similarity metrics can distinguish the trained prompts of the same tasks and different tasks, and (\romannumeral2) examine whether these metrics align with the zero-shot transfer performance.

% Regarding (\romannumeral1), we compare the similarities of the investigated metrics for prompts within the same task (trained with different random seeds) and between different tasks. As shown in Table \ref{table:similiarty_metrics}, we find that all the metrics can distinguish the prompts of the same task and different tasks well\footnote{The effectiveness of all metrics are left in \cref{ssec:appendix_effectiveness_of_similarity_metrics}.}, suggesting that the trained soft prompts of different tasks form distinguishable clusters in the embedding space and stimulate different abilities within the PLM.

Regarding (\romannumeral1), we compare the similarities of the investigated metrics for two trained prompts within the same task (trained with different random seeds) and between different tasks in Table~\ref{table:similiarty_metrics}. From the results, we can observe that all the metrics work well to distinguish the prompts of the same task and different tasks. This suggests that the trained soft prompts of different tasks form distinguishable clusters in the embedding space and also stimulate different abilities within the PLM. 

Moreover, to evaluate (\romannumeral2), how well the similarity metrics align with the cross-task transfer performance,
we quantify the correlations between the similarities and zero-shot transfer performance in Figure~\ref{fig:zero-shot-task-transfer}. Specifically, for each target task's prompt, we rank various source tasks' prompts with similarity scores and zero-shot transfer performance and then compute the Spearman's rank correlation~\citep{spearman1987proof} between the two ranks generated by these two ways. The overall results are shown in Table~\ref{table:prompt_similiarty_measure}\footnote{The detailed results by task types are left in \cref{ssec:appendix_correlation_between_prompt_transferability_and_prompt_similarity}.}.
%, and the detailed results by task types are left in \cref{ssec:appendix_correlation_between_prompt_transferability_and_prompt_similarity}. 
We can see that: (1) The \textit{overlapping rate of activated neurons} ($\mathrm{ON}$) metric works better than all the embedding similarities, which suggests that model stimulation is more important for prompt transferability than embedding distances. %(2) Among all the embedding similarity metrics, the two metrics based on Euclidean distances work poorly and even have negative correlations on averages. The metrics based on cosine similarities work better and the $\mathrm{C}_{\mathrm{average}}$ metric can achieve comparable performance to the $\mathrm{ON}$ metric using bottom layers. However, the ability of $\mathrm{C}_{\mathrm{average}}$ to distinguish different tasks (Table~\ref{table:similiarty_metrics}) is obviously worse. These results indicate that the properties of the prompt embedding space are tricky and hard to design transferable PT methods based on them. 
(2) $\mathrm{ON}$ works much worse on \TXXL ($11$B parameters) than on \RoBERTaLARGE ($330$M parameters). We guess this is because larger PLMs have higher redundancy~\citep{Aghajanyan2021IntrinsicDE}, which means prompts can activate different redundant neurons to do similar jobs and thus influence the sensitivity of $\mathrm{ON}$ metric. This is supported by the experiments showing that the Spearman's correlation scores of $\mathrm{ON}$ drop with the increase of PLM scales (Figure~\ref{fig:model_size_colrelation}), from which we can see $\mathrm{C}_{\mathrm{average}}$ also exhibits a similar trend. We encourage future work to explore how to overcome the PLM redundancy for better transferrable PT. As a preliminary trial, we find that by taking the intersection of activation states of $3$ prompts trained with different random seeds, $\mathrm{ON}$'s correlation score on \TXXL raises from $36.9\%$ to $46.3\%$.

\begin{table}[!t]
\begin{center} 
\small
%\begin{adjustbox}{max width=1\linewidth}
{
\setlength\tabcolsep{0.15em}

\begin{tabular}{c|l|cc}
\toprule
\textbf{Projector} & \textbf{Task} & \textbf{$\mathrm{C}_{\mathrm{average}}$} & \textbf{$\mathrm{ON}$} \\ 
\midrule
\multirow{3}{*}{\begin{tabular}[c]{@{}c@{}}Task Tuning\\ (\texttt{laptop})\end{tabular}} & \texttt{laptop} & 3.8 & 52.4 \\
& Same-Type Tasks & 4.1 & 51.0 \\
& Different-Type Tasks & 3.4  & 46.0 \\ \midrule
\multirow{3}{*}{\begin{tabular}[c]{@{}c@{}}Task Tuning\\ (\texttt{MNLI})\end{tabular}} & \texttt{MNLI} & 2.7 & 70.7 \\
& Same-Type Tasks & 2.7 & 56.7 \\
& Different-Type Tasks & 4.1 & 53.4 \\ \bottomrule
\end{tabular}

}
%\end{adjustbox}
\caption{Similarities (\%) between the prompts projected with \textbf{Task Tuning} projector and the original prompts trained on \TXXL.}
\label{table:original_prompt_and_projected_prompt_similiarty}
\end{center} 
\end{table}

We further explore whether the prompt similarity metrics also work in the cross-model transfer setting by testing whether they work between the projected prompts and original prompts of the same task. In Table~\ref{table:original_prompt_and_projected_prompt_similiarty}, we show the similarities of prompts projected with \textbf{Task Tuning} projectors by the two best metrics $\mathrm{C}_{\mathrm{average}}$ and $\mathrm{ON}$. We can see: (1) $\mathrm{ON}$ metric shows that the projected prompts are highly similar to the original prompts within the same type of projector-training tasks but are not so similar to different-type tasks, which is quite consistent with the cross-model zero-shot transfer performance in Table~\ref{table:robertalarge_to_t5xxl}. (2) However, $\mathrm{C}_{\mathrm{average}}$ cannot reflect this phenomenon, which shows that the perspective of model stimulation is more promising for understanding transferability again.

%% file: sections/conclusion.tex
\section{Conclusion}
\label{sec:conclusion}

We empirically investigate the transferability of prompts in this paper. In the cross-task setting, we find that soft prompts can transfer to similar tasks without training. In the cross-model setting, we successfully project prompts into the space of other PLMs. Further, we utilize trained prompts of other tasks or other PLMs as initialization to significantly accelerate training and improve effectiveness. Moreover, we explore various prompt transferability indicators and show that how the prompts stimulate PLMs is important to transferability. We hope the empirical analyses and the \textit{model stimulation} idea can facilitate further research on transferable and efficient PT.

%% file: sections/acknowledgments.tex
\section*{Author Contributions}
Yusheng Su and Xiaozhi Wang mainly initiated and organized this research. First, Xiaozhi Wang proposed the research idea. Yusheng Su, Xiaozhi Wang, Yujia Qin and Yankai Lin discussed most of the investigation settings. Yusheng Su, Chi-Min Chan, and Kaiyue Wen conducted the experiments. Xiaozhi Wang and Yusheng Su wrote the paper. Yankai Lin, Yujia Qin, Huadong Wang, Zhiyuan Liu, Peng Li, Juanzi Li, and Lei Hou revised the paper. Maosong Sun and Jie Zhou provided valuable advice and proofread this paper.

\section*{Acknowledgment}
\label{sec:acknowledgments}
This work is supported by the National Key Research and Development Program of China (No. 2020AAA0106500), Beijing Academy of Artificial Intelligence (BAAI), Institute Guo Qiang at Tsinghua University, and International Innovation Center of Tsinghua University, Shanghai China. We thank the anonymous reviewers for their valuable comments.

%% file: sections/appendix.tex
\clearpage

\appendix
\section{Basic Setup for Various Tasks}
\label{sec:task_setup}

\subsection{Dataset and Task}
\label{ssec:appendix_dataset_and_task}

\paragraph{Sentiment Analysis (SA)} SA is the task of classifying sentiment polarities for a given sentence. We select \texttt{IMDB} \cite{maas-EtAl:2011:ACL-HLT2011}, \texttt{SST-2} \cite{socher-etal-2013-recursive}, SemEval/\texttt{laptop} \cite{pontiki-etal-2014-semeval}, SemEval/\texttt{restaurant} \cite{pontiki-etal-2014-semeval}, Movie Rationales (\texttt{Movie}) \cite{zaidan-eisner-piatko-2008:nips}, and TweetEval (\texttt{Tweet}) \cite{barbieri-etal-2020-Tweeteval} for our experiments.

\paragraph{Natural Language Inference (NLI)} NLI is the task of determining whether a hypothesis is entailed or contradicted by a given sentences (premise, hypothesis). We select \texttt{MNLI} \citep{williams-etal-2018-broad}, \texttt{QNLI} \citep{wang2018glue}, and \texttt{SNLI} \citep{bowman-etal-2015-large} for our experiments.

\paragraph{Ethical Judgment (EJ)} EJ is the task of deciding whether a sentence is ethically acceptable. We select Ethics/\texttt{deontology} \cite{hendrycks2021aligning} and Ethics/\texttt{justice} \cite{hendrycks2021aligning} for our experiments.

\paragraph{Paraphrase Identification (PI)} PI is the task of classifying whether a pair of sentences are semantically identical. We select \texttt{QQP} \citep{sharma2019natural} and \texttt{MRPC} \citep{dolan-brockett-2005-automatically} for our experiments.

\paragraph{Question Answering~(QA)} QA is the task of answering a question. We choose \texttt{SQuAD} \citep{rajpurkar-etal-2016-squad} and \texttt{NQ-Open} \citep{lee-etal-2019-latent} to analyze. For \texttt{SQuAD}, a PLM captures the answer from the content. As for \texttt{NQ-Open}, a PLM directly generates the answer without the content.

\paragraph{Summarization~(SUM)} SUM is the task of summarizing a given article and generating the abstract. We select \texttt{Multi-News} \citep{fabbri-etal-2019-multi}, and \texttt{SAMSum} \citep{gliwa-etal-2019-samsum} for our experiments.

\subsection{Evaluation Metrics}
\label{ssec:appendix_evaluation_metric}
%For SA, NLI, EJ, and PI tasks, we choose accuracy (Acc.) as their evaluation metric to analyze. For QA and SUM tasks, we utilize F1 and ROUGE-L \cite{lin-2004-rouge}, respectively.
For classification tasks (SA, NLI, EJ, and PI), we use accuracy (Acc.) as their evaluation metric. As for generation tasks (QA and SUM), we utilize F1 and ROUGE-L \cite{lin-2004-rouge}, respectively.

\subsection{Prompt Tuning Setting}
\label{ssec:appendix_prompt_tuning_setting}
In the experiments, for all the investigated tasks, we use AdamW~\cite{loshchilov2018decoupled} as the optimizer and set the learning rate as $0.001$. We set the length of soft prompts $l$ as $100$. All the soft prompts are randomly initialized and optimized with Equation \ref{eq:continuous_prompt_tuning}. In the inference stage, RoBERTa predicts the label tokens at the $\mathtt{[MASK]}$ position and T5 directly uses its decoder to do generation. For the classification tasks (SA, NLI, EJ and PI), we obtaining answers in a ranking manner, i.e., we rank the label tokens by their likelihoods and regard the PLMs as predict the label of the label token with highest likelihood. For the conditional generation tasks (QA and SUM), we directly take the outputs of PLMs as their answers.

%for all investigated tasks,
\subsection{Label Tokens}
\label{ssec:appendix_verbalizers}
The used label tokens for the classification tasks (SA, NLI, EJ, PI) are shown in Table~\ref{table:task_verbalizers}. For generation tasks (QA, SUM), the desired output is just the annotated answers.

\begin{table}[!h]
\begin{center} 
%\small
\scalebox{0.725}
{
{
\setlength\tabcolsep{1em}
\begin{tabular}{l|c}
\toprule
  \textbf{Task} & \textbf{Label Tokens}  \\
  \midrule
  \multicolumn{2}{c}{Sentiment Analysis (SA)} \\
  \midrule
  {\texttt{IMDB}} & \texttt{{positive, negative}}\\
  {\texttt{SST-2}} & \texttt{{positive, negative}}\\
  {\texttt{laptop}} & \texttt{{positive, moderate, negative}}\\
  {\texttt{restaurant}} & \texttt{{positive, moderate, negative}}\\
  {\texttt{Movie}} & \texttt{{positive, negative}}\\
  {\texttt{Tweet}} & \texttt{{positive, moderate, negative}}\\
  \midrule
  \multicolumn{2}{c}{Natural Language Inference (NLI)} \\
  \midrule
  {\texttt{MNLI}} & \texttt{{yes, neutral, no}}\\
  {\texttt{QNLI}} & \texttt{{yes, no}}\\
  {\texttt{SNLI}} & \texttt{{yes, neutral, no}}\\
  \midrule
  \multicolumn{2}{c}{Ethical Judgment (EJ)} \\
  \midrule
  {\texttt{deontology}} & \texttt{{acceptable, un}}\\
  {\texttt{justice}} & \texttt{{acceptable, un}}\\
  \midrule
  \multicolumn{2}{c}{Paraphrase Identification (PI)}  \\
  \midrule
  {\texttt{QQP}} & \texttt{{true, false}}\\
  {\texttt{MRPC}} & \texttt{{true, false}}\\

\bottomrule
\end{tabular}}}
\caption{Label tokens of classification tasks.}
\label{table:task_verbalizers}
\end{center} 
\end{table}

\begin{figure*}[!htbp]
    \centering
    \subfigure[\TSMALL]{
	    \includegraphics[width=0.482\textwidth]{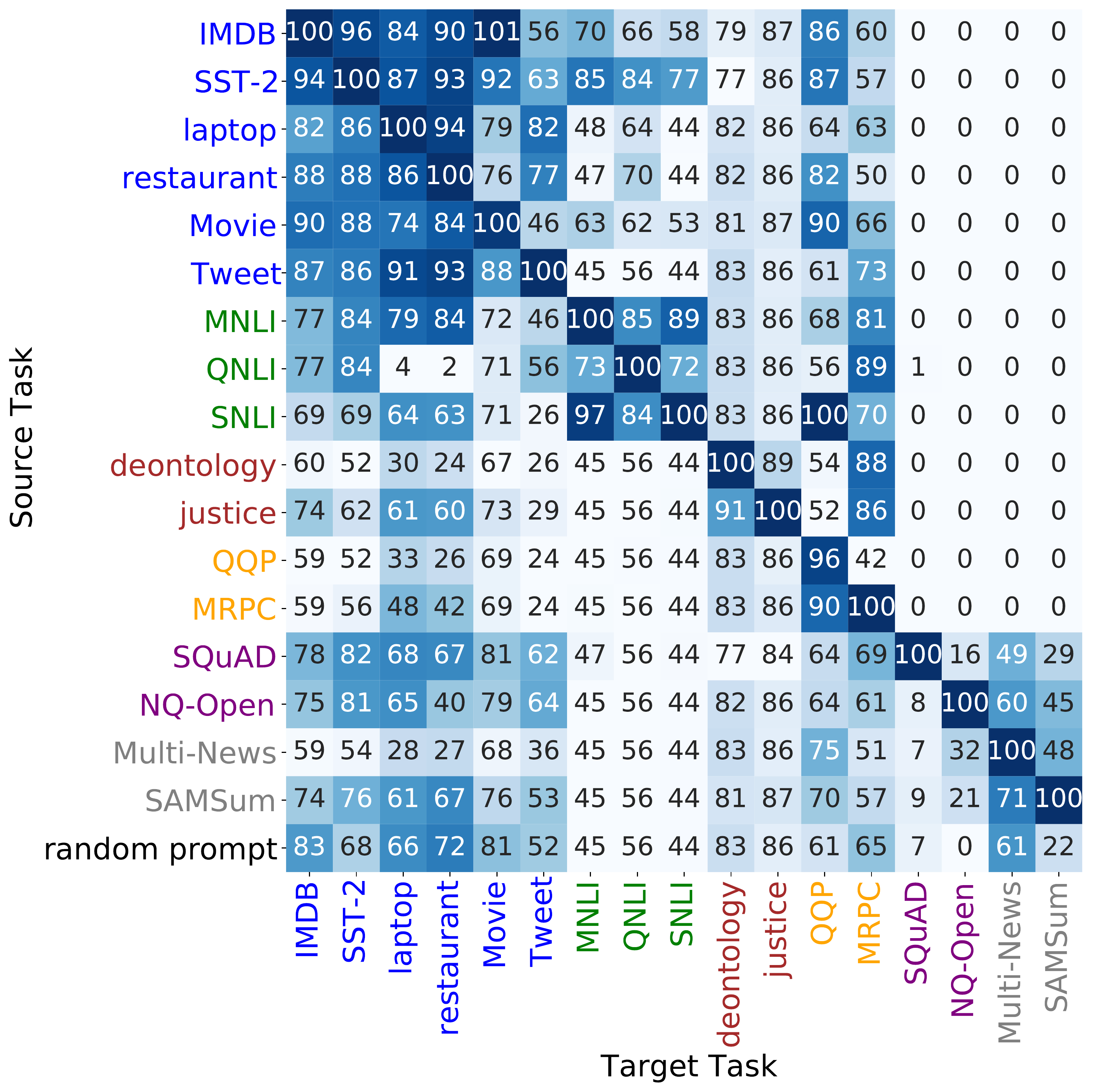}
    }
    \subfigure[\TBASE]{
	    \includegraphics[width=0.482\textwidth]{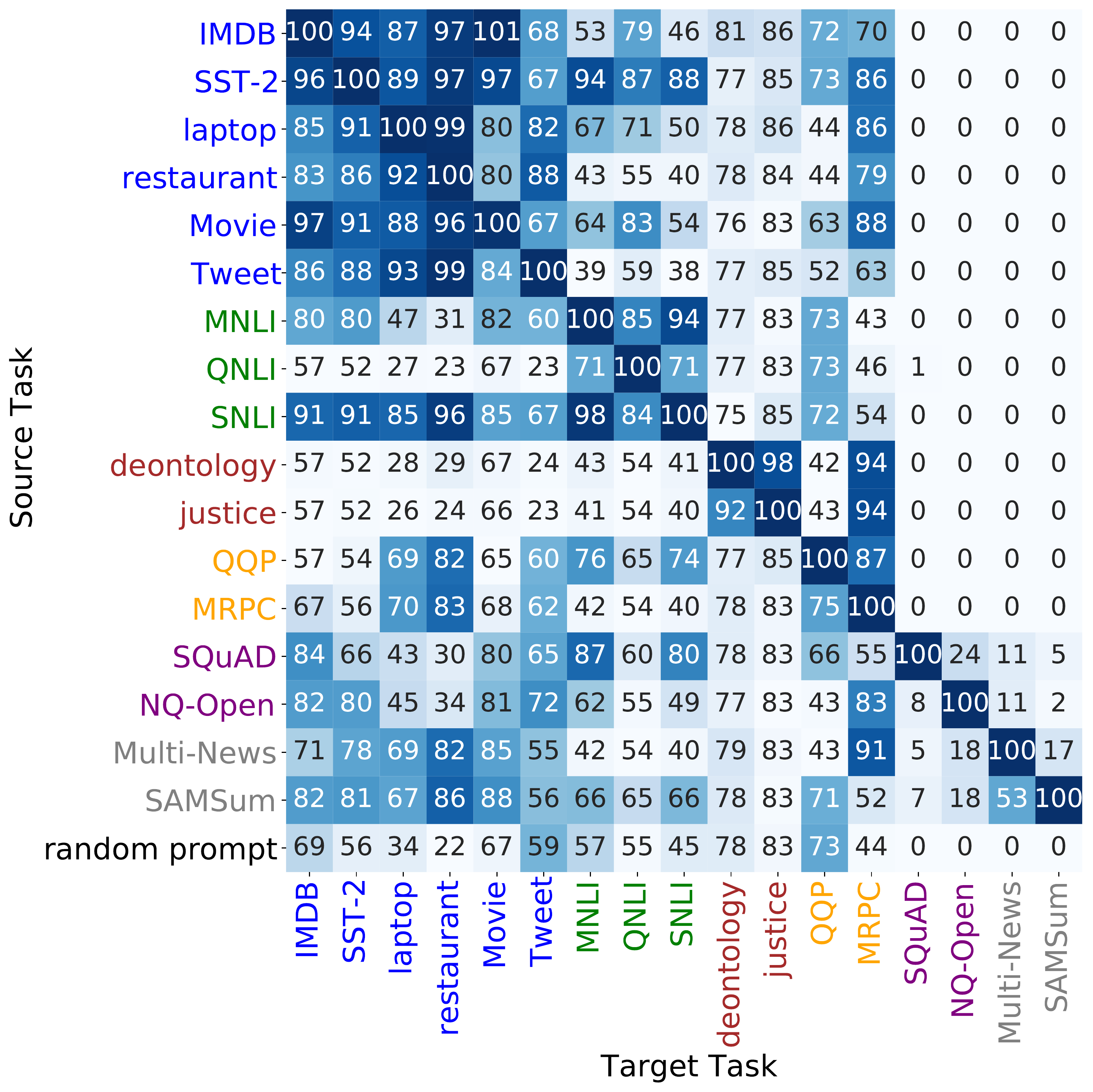}
    }
    % \subfigure[\TXXL]{
	   % \includegraphics[width=0.482\textwidth]{figs/task_transferability_2_results_t5_xxl.pdf}
    % }
    \subfigure[\RoBERTaBASE]{
	    \includegraphics[width=0.482\textwidth]{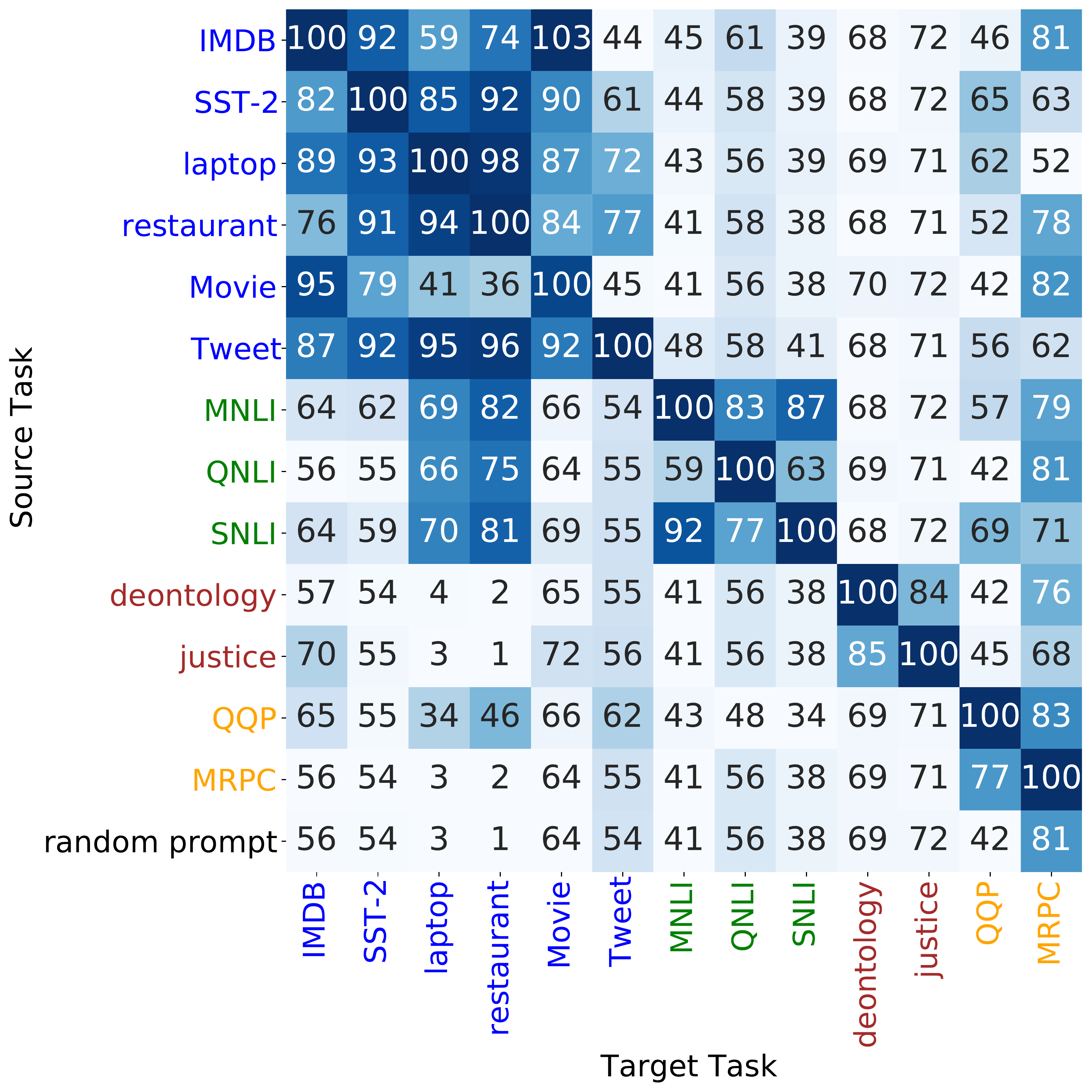}
    }
    \subfigure[\RoBERTaLARGE]{
	    \includegraphics[width=0.482\textwidth]{figs/task_transferability_2_results_roberta_large.pdf}
    }
    \caption{Relative performance (transferring zero-shot performance / original PT performance) (\%) on the target tasks (columns) of the soft prompts trained on the source tasks (rows), both of which demonstrate the relative performance for zero-shot transfer of prompts of RoBERTa and T5. Colors of the tasks names indicate the task types. \textbf{Blue}: sentiment analysis (SA). \textcolor{OliveGreen}{Green}: natural language inference (NLI). \textcolor{Brown}{Brown}: ethical judgment (EJ). \textcolor{YellowOrange}{Orange}: paraphrase identification (PI). \textcolor{Purple}{Purple}: question answering (QA). \textcolor{Gray}{Gray}: summarization (SUM). \textit{Random Prompt} of the last row means the soft prompts are randomly generated without any training.}
    \label{fig:appendix_zero-shot-task-transfer_1}
\end{figure*}

\begin{figure*}[!htbp]
    \centering
    \subfigure[Directly transferring (\RoBERTaBASE)]{
	    \includegraphics[width=0.482\textwidth]{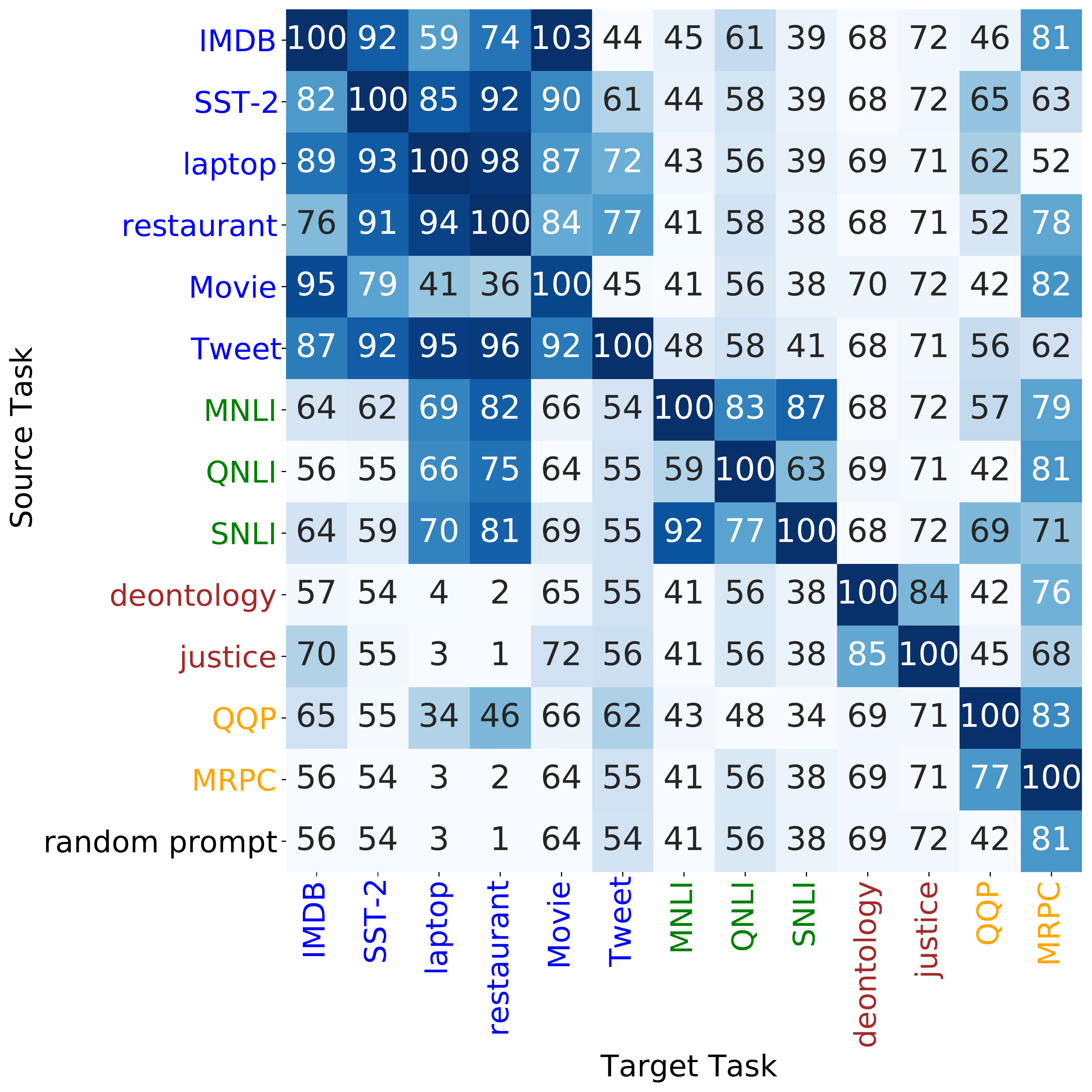}
    }
    \subfigure[Unifying the label tokens (\RoBERTaBASE)]{
	    \includegraphics[width=0.482\textwidth]{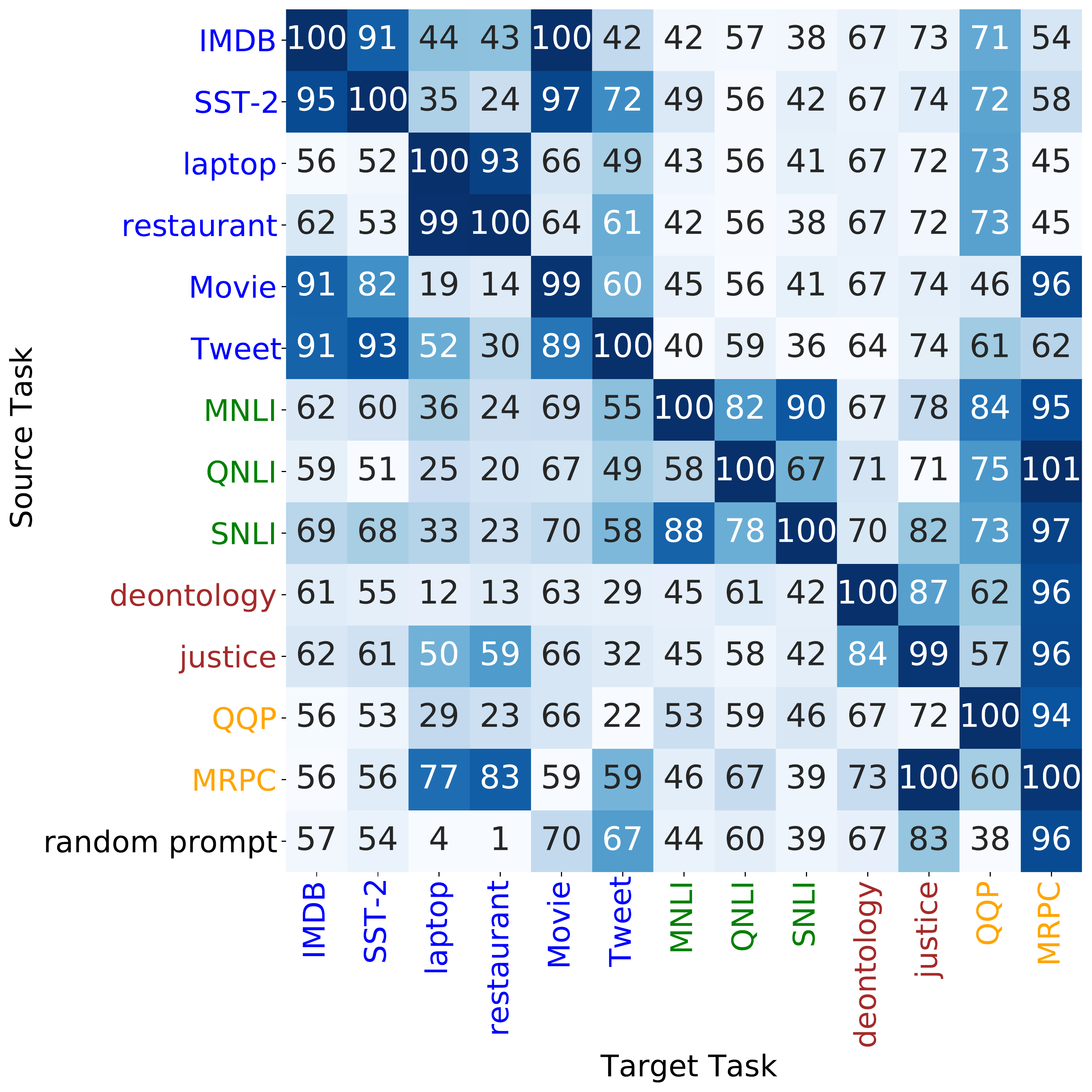}
    }
    \caption{To exclude the poor transferability, which may result from the fact that different-type tasks use different label tokens, we unify the label tokens of different tasks into the same set of numbers (\texttt{1}, \texttt{2}, $\ldots$) and choose \RoBERTaBASE for the experiments. From Figure (a) and (b), we observe that the transferability between different-type tasks are still generally not improved in this way. This indicates that different-type tasks surely require distinct abilities.}
    \label{fig:appendix_zero-shot-task-transfer}
\end{figure*}

\section{Cross-Task Transfer}
\label{ssec:appendix_cross-task_transfer}
\subsection{More Zero-shot transfer performance}
\label{ssec:appendix_relative_performance_of_zero-shot_transferability_on_various_plms}
In \cref{ssec:cross_task_performance}, we report the zero-shot transfer performance (relative performance) on \RoBERTaLARGE and \TXXL. Here, we investigate the zero-shot transfer performance on other sizes of RoBERTa and T5, which are shown in Figure~\ref{fig:appendix_zero-shot-task-transfer_1}. According to these results, we can find that the transferability of soft prompts between the tasks of different types is generally poor, which is consistent with the conclusion in \cref{ssec:cross_task_performance}.

\subsection{Unifying Label Tokens}
\label{ssec:further_observing_prompt_transferability_with_unifying_the_verbalizers}
% In \cref{ssec:cross_task_performance}, we find that the transferability of soft prompts between the tasks of different types is generally poor, and transferring soft prompts only achieve similar performances to the given randomly initialized prompt in many cases. 
We hypothesize that the poor transferability between different task types may result from the fact that different-type tasks usually use different label tokens, e.g., \texttt{yes} and \texttt{no} are for NLI tasks while \texttt{positive} and \texttt{negative} are for SA tasks. To verify whether this factor influences the transferability, we unify the label tokens of different tasks into the same set of numbers (\texttt{1}, \texttt{2}, $\ldots$) and choose \RoBERTaBASE for the experiments. In Figure~\ref{fig:appendix_zero-shot-task-transfer}, we can observe that the transferability between different-type tasks are generally not improved in this way. This indicates that different-type tasks surely require distinct abilities, which prohibits reusing prompts between them.

\subsection{Speedup Calculation}
\label{ssec:appendix_details_of_tpttask_speedup}

\begin{table*}[!t]
\begin{center} 
%\small
\begin{adjustbox}{max width=1\linewidth}
{
\setlength\tabcolsep{0.2em}
\begin{tabular}{l|l|*{6}{d{3.2}}|*{3}{d{3.2}}|*{2}{d{3.2}}|*{2}{d{3.2}}}
% \begin{table*}[!t]
% \begin{center}
% \begin{adjustbox}{max width=1\linewidth}
% {
% \setlength\tabcolsep{0.2em}
% \begin{tabular}{c|cl|*{6}{d{3.2}}|*{3}{d{3.2}}|*{2}{d{3.2}}|*{2}{d{3.2}}}
\toprule
\multicolumn{2}{c|}{\textbf{Method}} &
\multicolumn{6}{c|}{\textbf{SA}} & \multicolumn{3}{c|}{\textbf{NLI}} & \multicolumn{2}{c|}{\textbf{EJ}} & \multicolumn{2}{c}{\textbf{PI}} \\
\cline{3-15} 
\multicolumn{2}{c|}{} & \mc{IMDB} & \mc{SST-2} & \mc{laptop} & \mc{restaurant} & \mc{Movie} & \mcl{Tweet} & \mc{MNLI} & \mc{QNLI} & \mcl{SNLI} & \mc{deontology} & \mcl{justice} & \mc{QQP} & \mc{MRPC}\\ 
\bottomrule
\multicolumn{15}{c}{From \BertBASE to \RoBERTaBASE} \\
\midrule
\multicolumn{2}{l|}{PT on \RoBERTaBASE} & \mc{{89.9}} & \mc{{93.8}} & \mc{{77.3}} & \mc{{80.7}} & \mc{{79.2}} & \mcl{{74.5}} & \mc{{80.6}} & \mc{{90.5}} & \mcl{{88.5}} & \mc{{72.9}} & \mcl{{70.0}} & \mc{{86.9}} & \mc{{83.9}} \\
\midrule 
\multicolumn{2}{l|}{Random Prompt} & 50.6 & 50.8 & 2.3 & 1.2 & 50.5 & 40.5 & 32.8 & 50.5 & 33.3 & 50.4 & 50.2 & 36.8 & 68.0 \\
\midrule
\multirow{2}{*}{IMDB, laptop} & Distance Minimizing  & \mc{\textbf{89.7}} & 53.1 & \mc{\textbf{75.6}} & 18.3 & 54.2 & 24.0 & 31.2 & 50.0 & 33.3 & 50.6 & 50.0 & 36.8 & 67.2 \\
 & Task Tuning & \mc{\textbf{88.2}} & \mc{\textbf{82.2}} & \mc{\textbf{76.3}} & \mc{\textbf{77.9}} & \mc{\textbf{73.4}} & 43.6 & 32.0 & 47.9 & 32.8 & 49.8 & 49.4 & 50.2 & 47.7\\
\midrule
\multicolumn{1}{l|}{\multirow{2}{*}{MNLI}} & Distance Minimizing & 55.6 & 51.0 & 2.5 & 1.4 & 53.1 & 41.1 & \mc{\textbf{80.0}} & 50.6 & 33.3 & 50.6 & 50.0 & 48.3 & 68.0 \\
 & Task Tuning & 50.9 & 52.0 & 11.9 & 13.1 & 45.8 & 18.2 & \mc{\textbf{80.0}} & \mc{\textbf{74.9}} & \mcl{\textbf{80.0}} & 50.4 & 49.9 & 36.8 & 68.1 \\ 
\bottomrule
\multicolumn{15}{c}{From \RoBERTaBASE to \RoBERTaLARGE} \\
\midrule
\multicolumn{2}{l|}{PT on \RoBERTaLARGE} & \mc{{91.8}} & \mc{{96.0}} & \mc{{78.1}} & \mc{{81.7}} & \mc{{81.7}} & \mcl{{76.6}} & \mc{{88.5}} & \mc{{93.4}} & \mcl{{90.7}} & \mc{{85.6}} & \mcl{{81.1}} & \mc{{89.0}} & \mc{{82.7}} \\
\midrule 
\multicolumn{2}{l|}{Random Prompt} & 50.1 & 50.2 & 2.0 & 2.0 & 49.5 & 40.5 & 32.7 & 51.0 & 33.3 & 50.3 & 49.9 & 40.6 & 61.2 \\ 
\midrule 
\multirow{2}{*}{IMDB, laptop} & Distance Minimizing & \mc{\textbf{92.1}} & 50.1 & \mc{\textbf{77.0}} & 1.4 & 51.0 & 37.6 & 33.1 & 50.2 &  32.8 & 50.4 & 50.0 & 62.3 & 38.3 \\
 & Task Tuning & \mc{\textbf{90.4}} & \mc{\textbf{76.2}} & \mc{\textbf{64.2}} & \mc{\textbf{69.5}} & \mc{\textbf{79.7}} & 45.0 & 33.3 & 50.5 & 33.1  & 50.3 & 50.0 & 38.5 & 79.7\\
\midrule 
\multirow{2}{*}{MNLI} & Distance Minimizing & 50.3 & 51.2 & 5.2 & 5.9 & 51.0 & 40.6 & \mc{\textbf{88.5}} & 49.1 & 33.2 & 50.3 & 50.0 & 45.1 & 66.4 \\
 & Task Tuning & 67.7 & 76.1 & 28.9 & 43.7 & 60.4 & 49.1 & \mc{\textbf{87.1}} & \mc{\textbf{79.4}} & \mcl{\textbf{84.5}} & 49.7 & 50.0 & 36.8 & 68.5 \\
\bottomrule
\multicolumn{15}{c}{From \TBASE to \TXXL} \\
\midrule
\multicolumn{2}{l|}{PT on \TXXL} & \mc{{96.5}} & \mc{{97.4}} & \mc{{76.6}} & \mc{{88.1}} & \mc{{97.9}} & \mcl{{72.5}} & \mc{{90.5}} & \mc{{95.2}} & \mcl{{93.4}} & \mc{{87.0}} & \mcl{{92.5}} & \mc{{90.0}} & \mc{{86.3}}\\
\midrule 
\multicolumn{2}{l|}{Random Prompt} & 49.7 & 49.0 & 19.8 & 17.0 & 51.6 & 15.5 & 31.8 & 49.3 & 31.9 & 51.3 & 50.0 & 36.4 & 67.0 \\
\midrule 
\multirow{2}{*}{laptop} & Distance Minimizing & 49.0 & 49.7 & \mc{\textbf{76.6}} & 17.0 & 52.3 & 16.3 & 31.8 & 48.7 & 33.3 & 54.1 & 49.0 & 36.7 & 67.7  \\
 & Task Tuning & \mc{\textbf{77.2}} & \mc{\textbf{86.2}} & \mc{\textbf{80.3}} & \mc{\textbf{83.5}} & \mc{\textbf{64.6}} & \mcl{\textbf{55.2}} & 31.9 & 49.9 & 32.9 & 48.7 & 52.8 & 50.7 & 53.1  \\
\midrule 
\multirow{2}{*}{MNLI} & Distance Minimizing & 49.7 & 49.0 & 19.8 & 17.1 & 51.6 & 15.5 & \mc{\textbf{90.5}} & 49.3 & 34.8 & 52.3 & 50.0 & 36.8 & 67.7 \\
 & Task Tuning & 54.9 & 70.0 & 60.8 & 74.1 & 3.6 & 41.4 & 
\mc{\textbf{89.7}} & \mc{\textbf{84.8}} & \mcl{\textbf{90.8}} & 49.7 & 50.0 & 37.2 & 66.4  \\
\bottomrule
\end{tabular}
}
\end{adjustbox}
\caption{We conduct experiments between various PLMs in different scales and heterogeneous frameworks: from \BertBASE to \RoBERTaBASE, from \RoBERTaBASE to \RoBERTaLARGE, and from \TBASE to \TXXL. Besides, we highlight the non-trivial zero-shot performance (\%) of the cross-model setting with \textbf{bold}.}
\label{table:appendix_various_plms_to_various_plms}
\end{center} 
\end{table*}

In this paper, we compute convergence speedup and comparable-result speedup as follows:
\begin{equation}
    \small
    \label{eq:appendix_convergence_speedup}
    \begin{aligned}
        \begin{gathered}
        \text{Convergence Speedup} (\mathrm{x}) = \frac{\text{PT convergence time}}{\text{TPT convergence time}}, \\ \\
        \text{Comparable-result Speedup} (\mathrm{x}) = \\
        \frac{\text{PT convergence time}}{\text{time of TPT achieving  comparable result to PT}}.
        \end{gathered}
    \end{aligned}
\end{equation}

%As shown in Figure～\ref{fig:appendix_initialization_experiments_cross-task}, 
We calculate the training loss and the evaluation score per $100$ steps during the training. When the training loss stops dropping and the evaluation score stops increasing for $300$ steps, we set the point as the convergence point. For the convergence speedup in Equation \ref{eq:appendix_convergence_speedup}, the PT convergence time is divided by the TPT convergence time. As for the comparable-result speedup in Equation \ref{eq:appendix_convergence_speedup}, the PT convergence time are divided by the time of TPT achieving comparable performance to PT.

\section{Cross-Model Transfer}
\label{ssec:appendix_cross-model_transfer}

\subsection{Implementation Details of Projector}
\label{ssec:appendix_implementation_details_of_projectors}
As mentioned in \cref{ssec:cross-model_projector}, we give the prompt of the source PLM, $P^s=\{\mathbf{p}_{1}^{s},\ldots,\mathbf{p}_{l}^{s}\}$, and concatenate its $l$ virtual tokens into a unified vector $\mathbf{P}^{s}\in\mathbb{R}^{l d_s}$, where $d_s$ is the hidden size of the source PLM. To transfer $\mathbf{P}^{s}$ to the target PLM whose hidden size is $d_t$, we design a projection function $\mathbf{Proj}(\cdot)$ parameterized by a two-layer perceptron as follows:
\begin{equation}
    \mathbf{\tilde{P}}^{s}\! =\! \mathbf{Proj}(\mathbf{P}^{s})\! =\! \mathbf{W}_{2}(\sigma(\mathbf{P}^{s}\mathbf{W}_{1}\!+\!\mathbf{b}_{1}))\!+\!\mathbf{b}_{2},
\end{equation}
where $\mathbf{W}_{1}\in\mathbb{R}^{d_h\times l d_s},\mathbf{W}_{2}\in \mathbb{R}^{l d_t \times d_h}$ are trainable matrices, $\mathbf{b}_1\in \mathbb{R}^{d_h},\mathbf{b}_2 \in \mathbb{R}^{l d_t}$ are biases, $\sigma$ is a non-linear activation function. For training configurations of projector, the optimizer is AdamW~\cite{loshchilov2018decoupled}, the training batch size is $16$, the learning rate is $0.005$, and the inner hidden size $d_h$ is $768$. In this paper, we investigate cross-model transfer among various PLMs including \BertBASE, \RoBERTaBASE, \RoBERTaLARGE, \TSMALL, \TBASE, and \TXXL, whose hidden sizes are $768$, $768$, $1024$, $512$, $768$, and $1024$, respectively. Besides, for non-linear activation functions, we have tried \texttt{tanh} and \texttt{LeakyReLU}~\cite{xu2015empirical}, and find their performance on various PLMs are similar. The reported results are based on \texttt{LeakyReLU}.

\begin{figure*}[!th]
\subcapraggedrighttrue
\subcaphangtrue
    \centering
    
    \subfigure[MNLI training loss. \texttt{[Without LayerNorm]}]{
	    \includegraphics[width=0.225\textwidth]{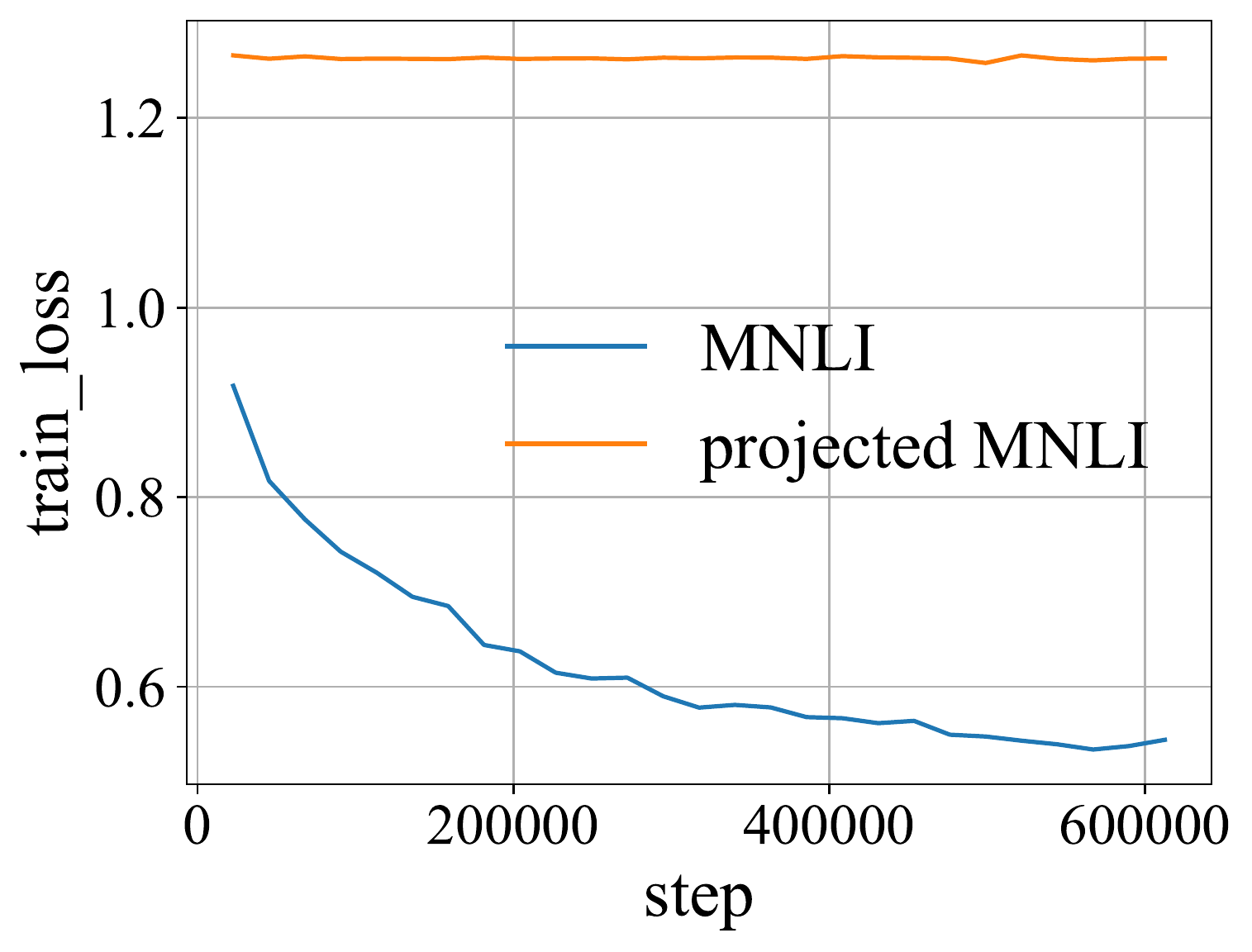}
    }
    \subfigure[MNLI evaluation accuracy. ~~~~\texttt{[Without LayerNorm]}]{
	    \includegraphics[width=0.225\textwidth]{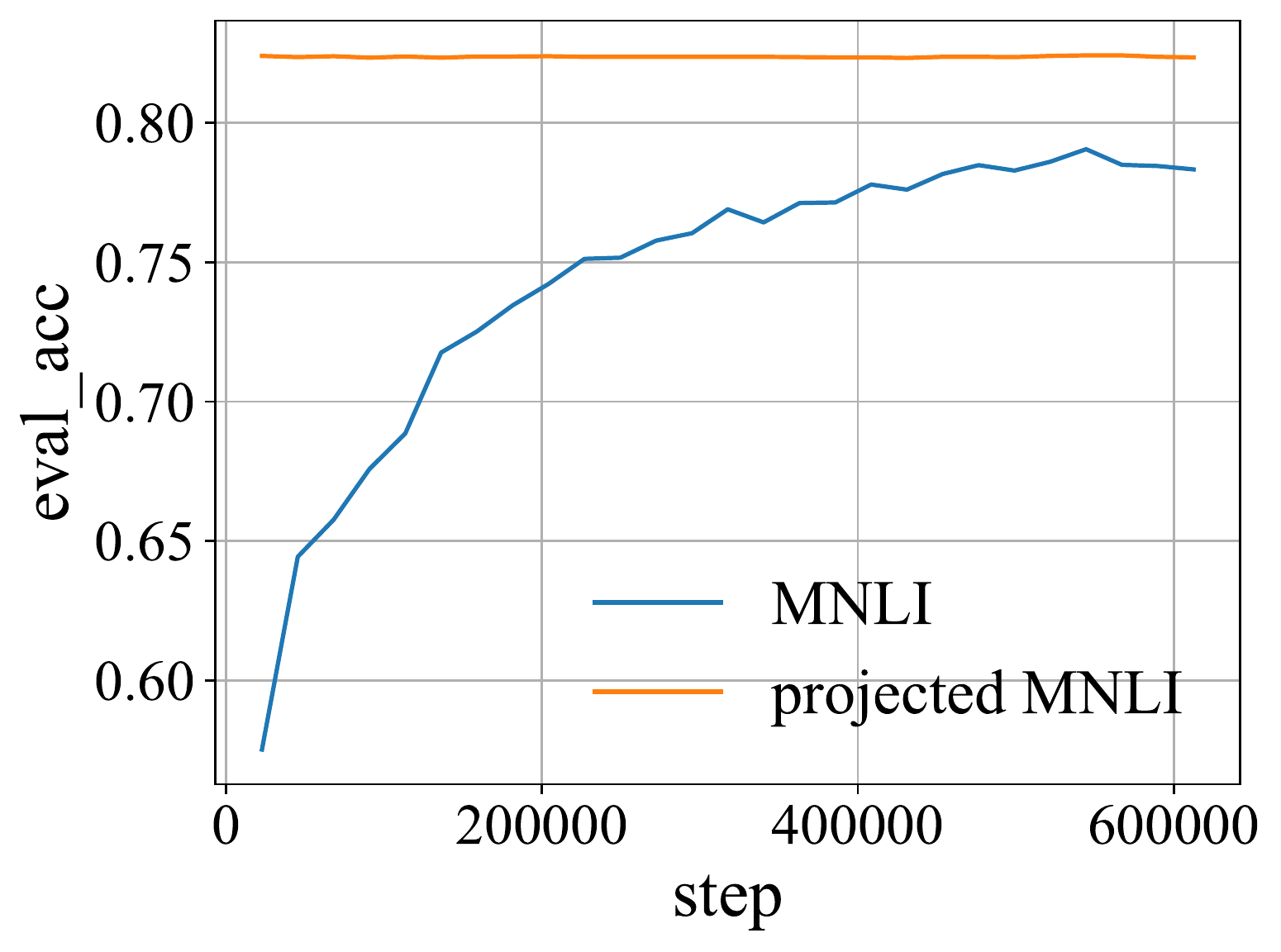}
    }
    \subfigure[MNLI training loss. \texttt{[With LayerNorm]}]{
	    \includegraphics[width=0.225\textwidth]{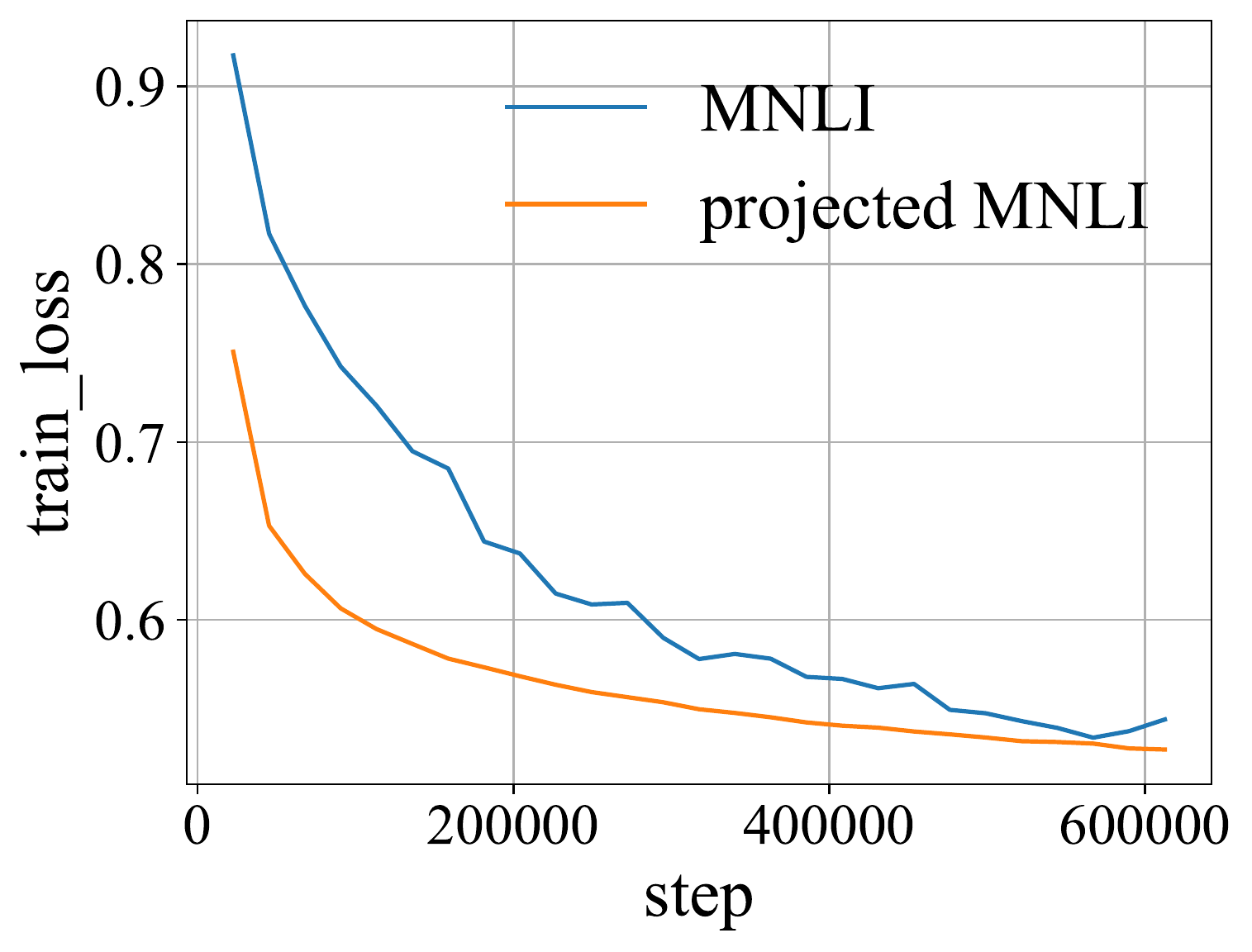}
    }
    \subfigure[MNLI evaluation accuracy. \texttt{[With LayerNorm]}]{
	    \includegraphics[width=0.225\textwidth]{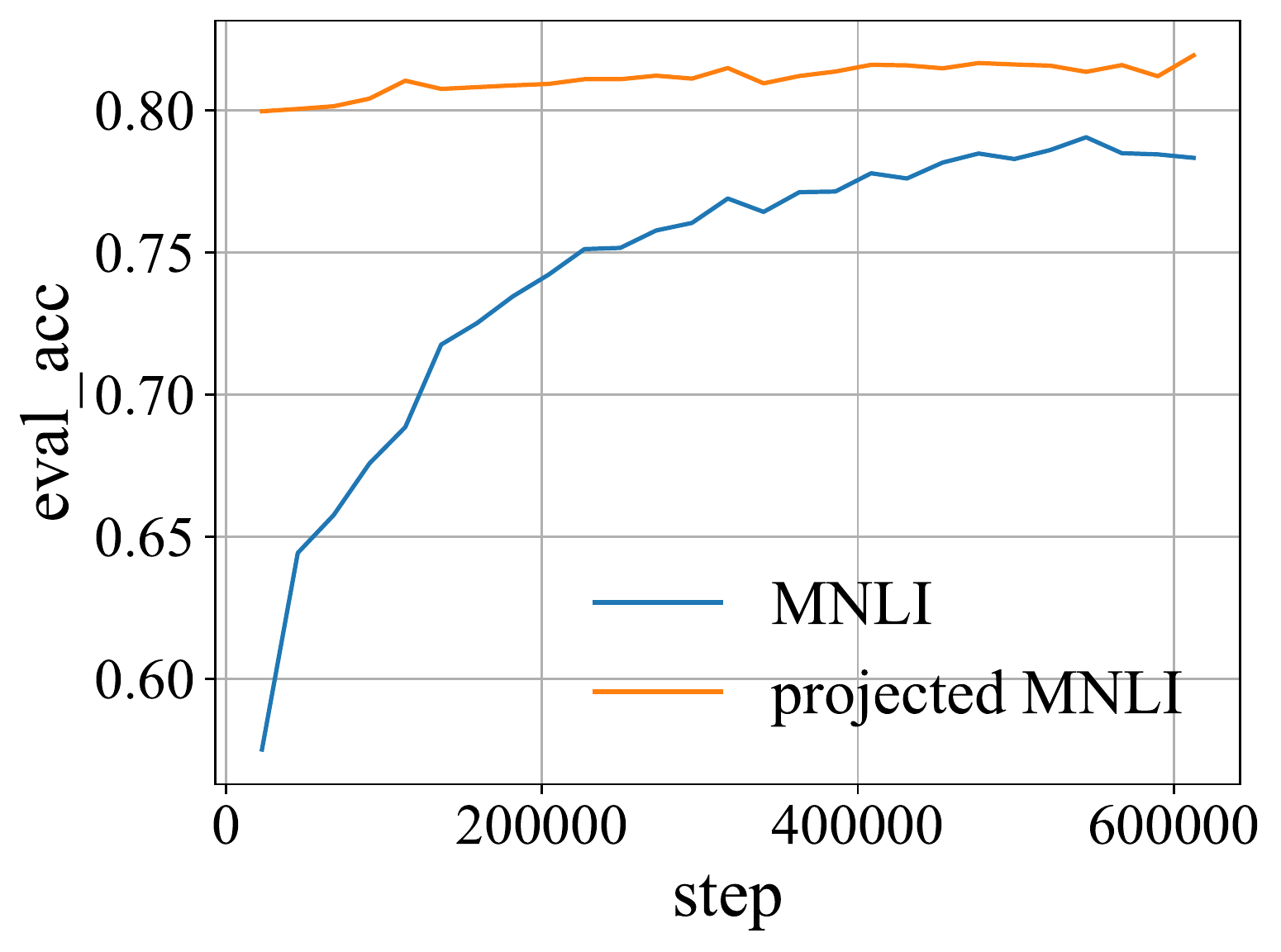}
    }

    \subfigure[IMDB training loss. \texttt{[Without LayerNorm]}]{
	    \includegraphics[width=0.225\textwidth]{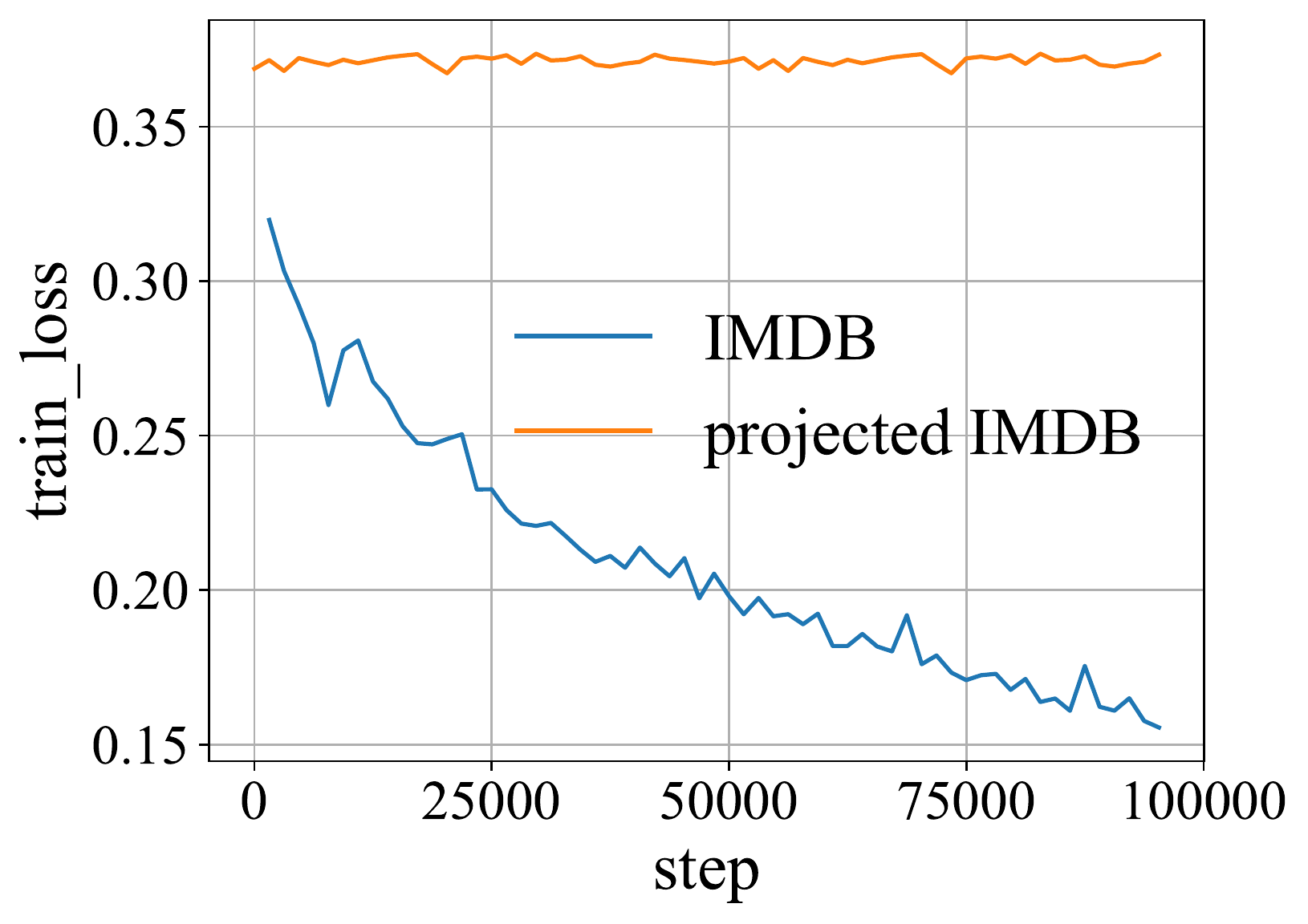}
    }
    \subfigure[IMDB evaluation ~accuracy. \texttt{[Without LayerNorm]}]{
	    \includegraphics[width=0.225\textwidth]{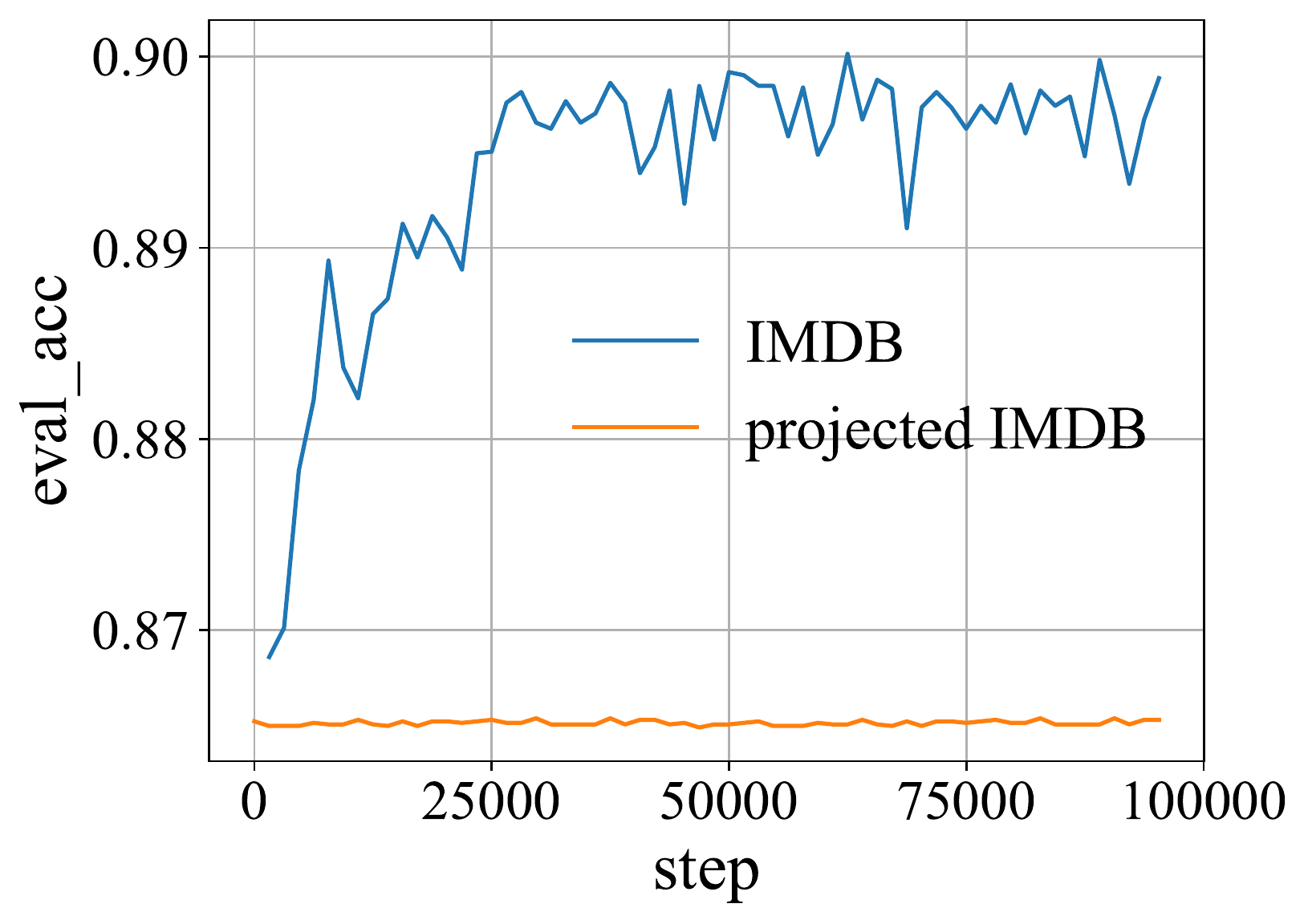}
    }
    \subfigure[IMDB training loss. \texttt{[With LayerNorm]}]{
	    \includegraphics[width=0.225\textwidth]{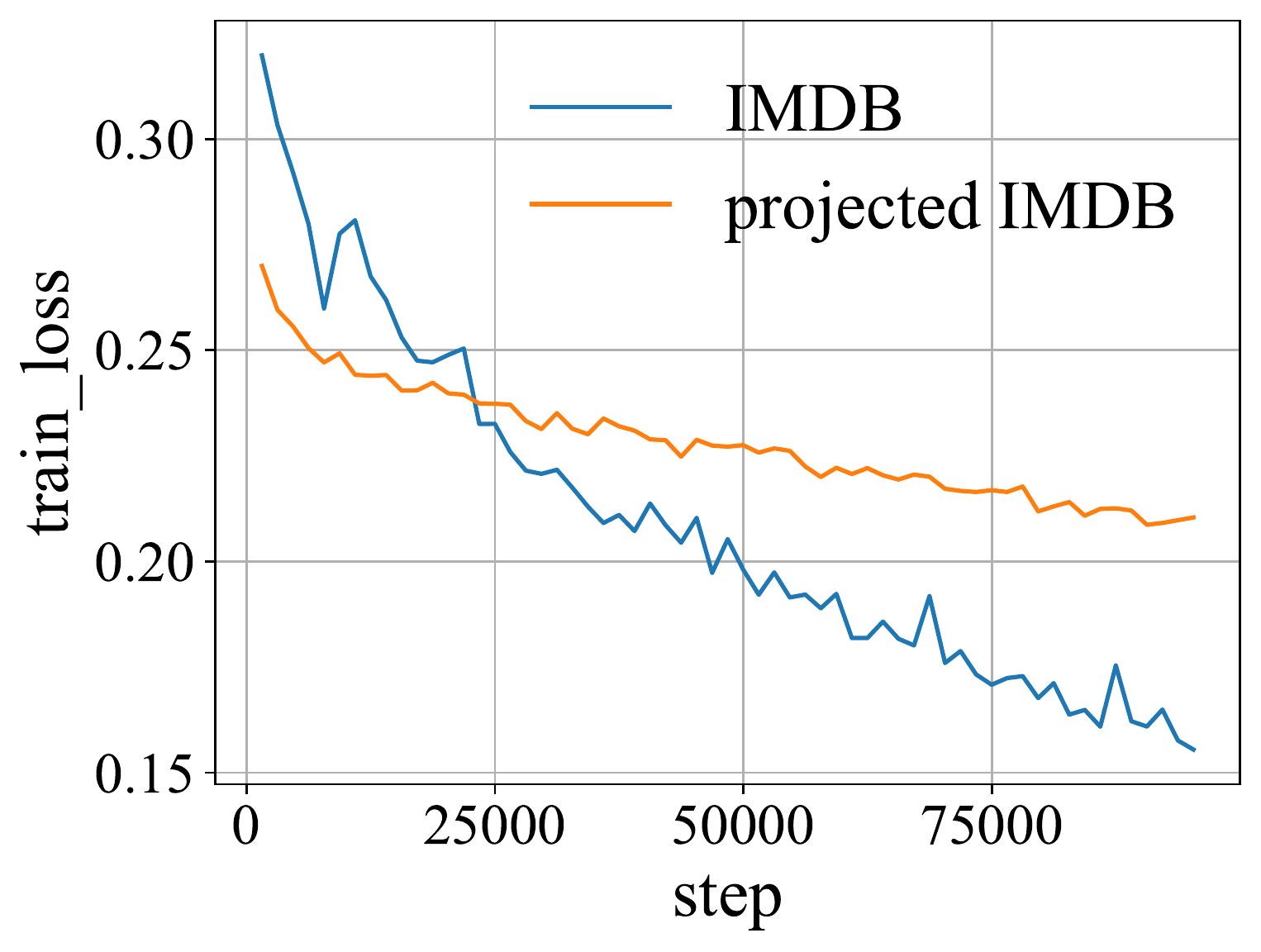}
    }
    \subfigure[IMDB evaluation accuracy. \texttt{[With LayerNorm]}]{
	    \includegraphics[width=0.225\textwidth]{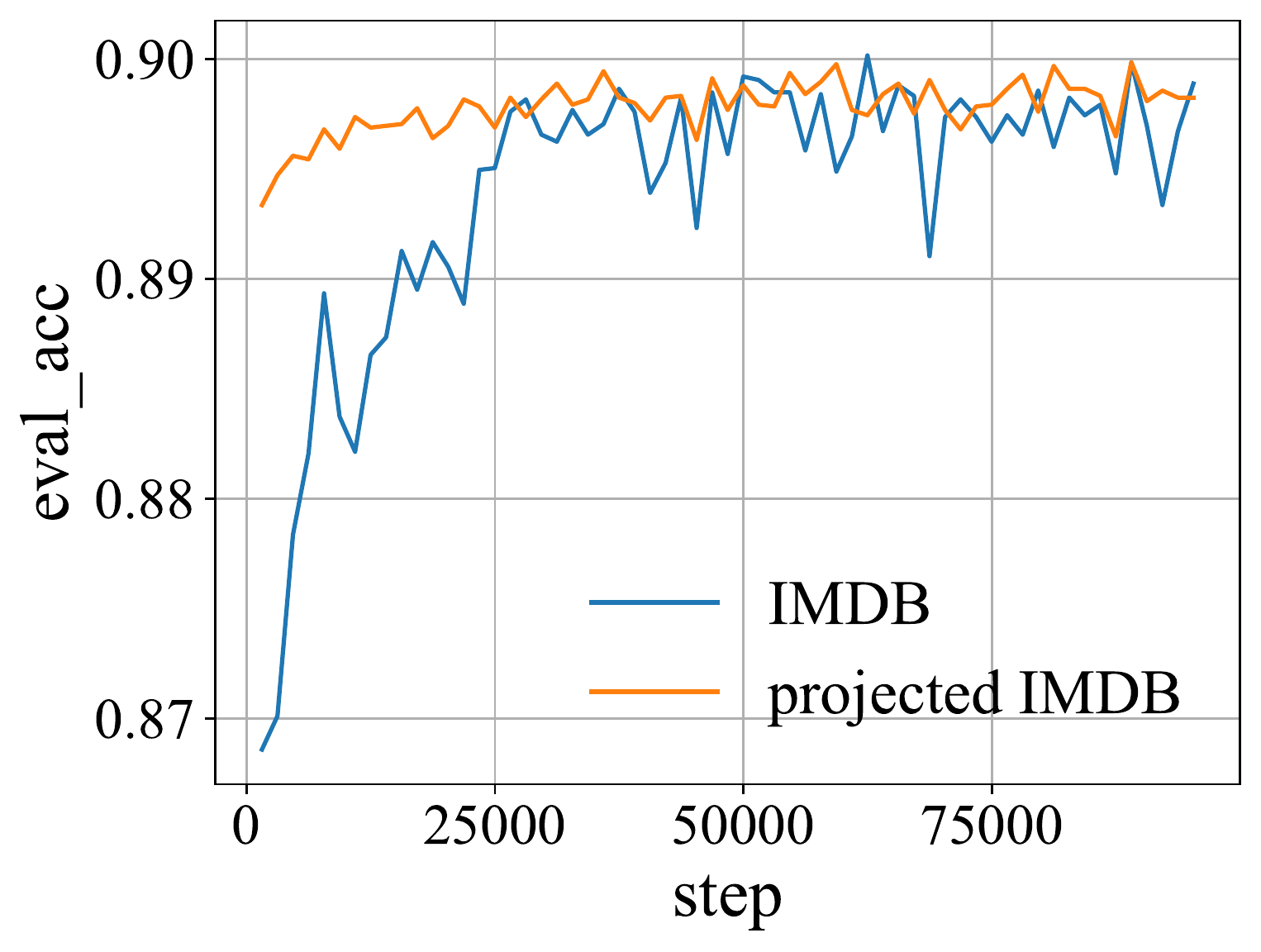}
    }

    \subfigure[restaurant training loss. \texttt{[Without LayerNorm]}]{
	    \includegraphics[width=0.225\textwidth]{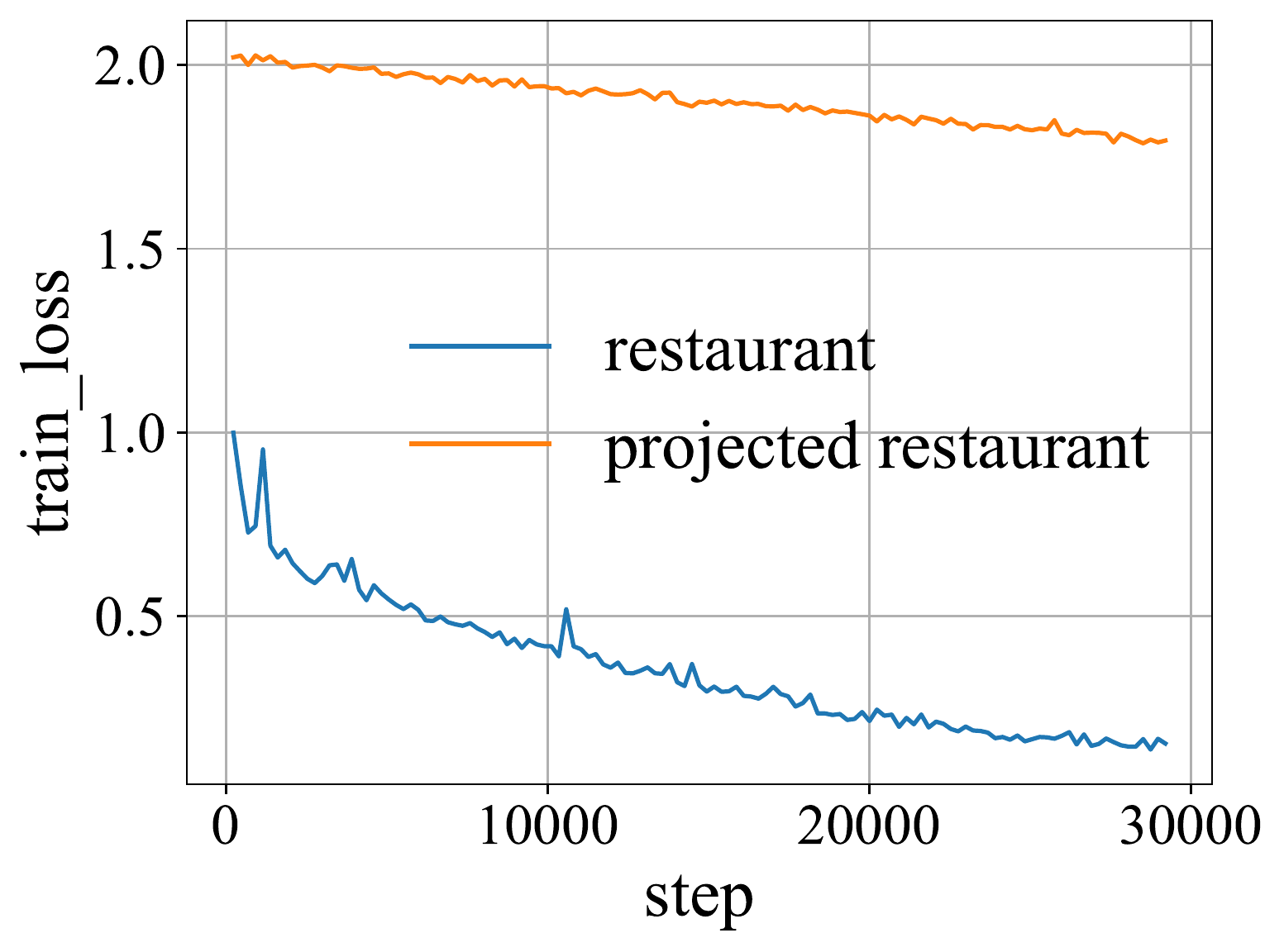}
    }
    \subfigure[restaurant evaluation accuracy. \texttt{[Without LayerNorm]}]{
	    \includegraphics[width=0.225\textwidth]{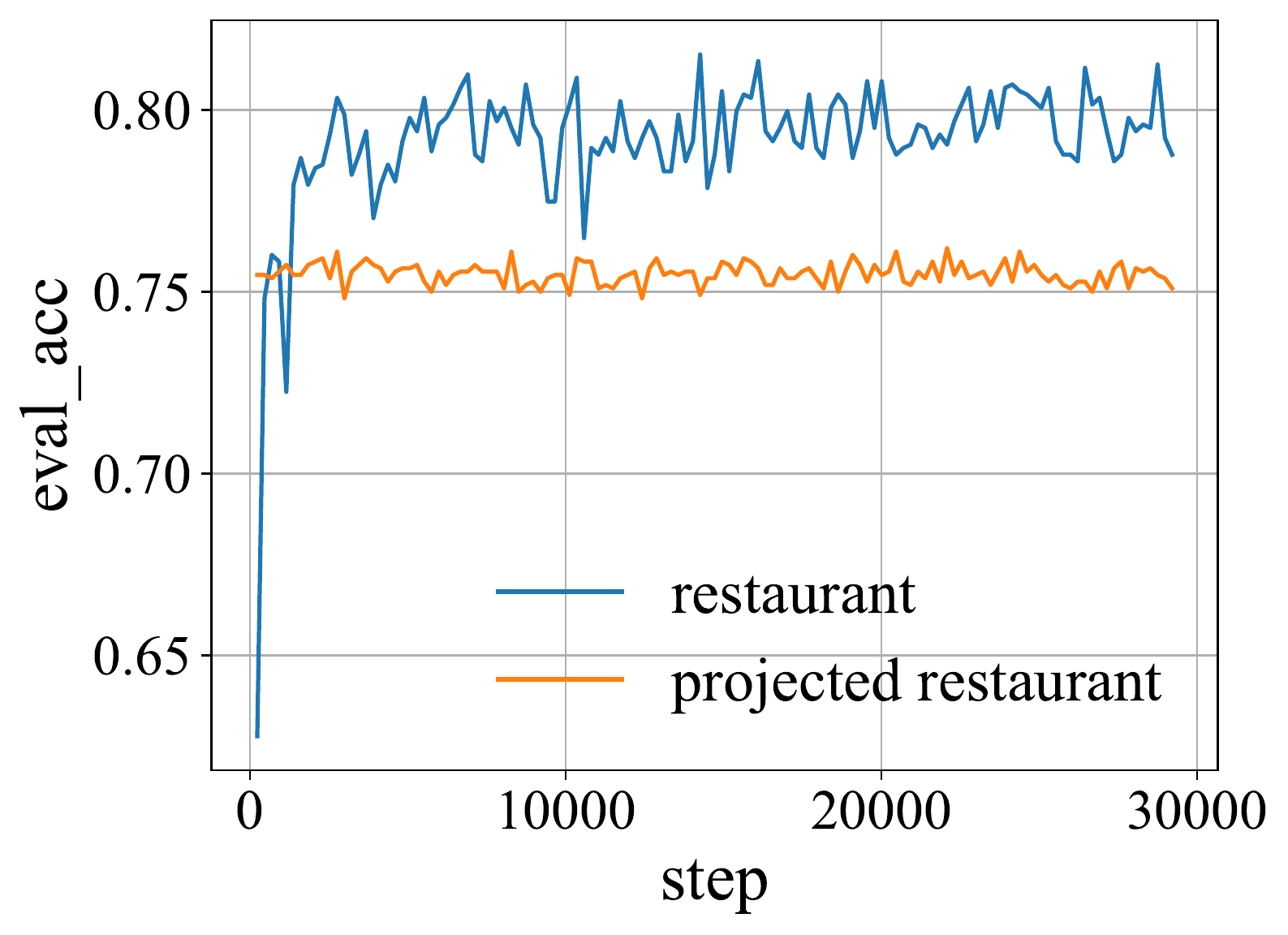}
    }
    \subfigure[restaurant training loss. \texttt{[With LayerNorm]}]{
	    \includegraphics[width=0.225\textwidth]{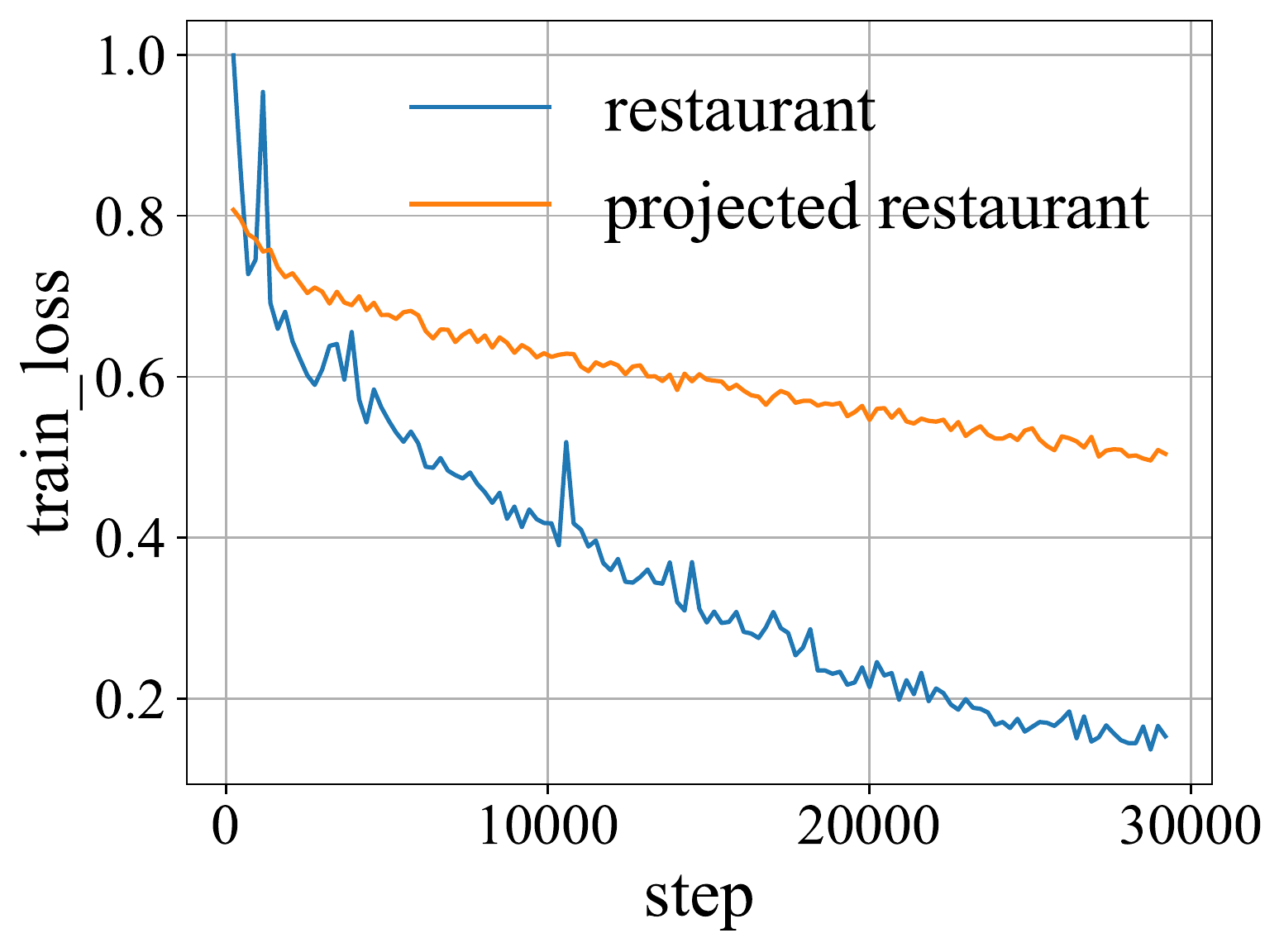}
    }
    \subfigure[restaurant evaluation accuracy. \texttt{[With LayerNorm]}]{
	    \includegraphics[width=0.225\textwidth]{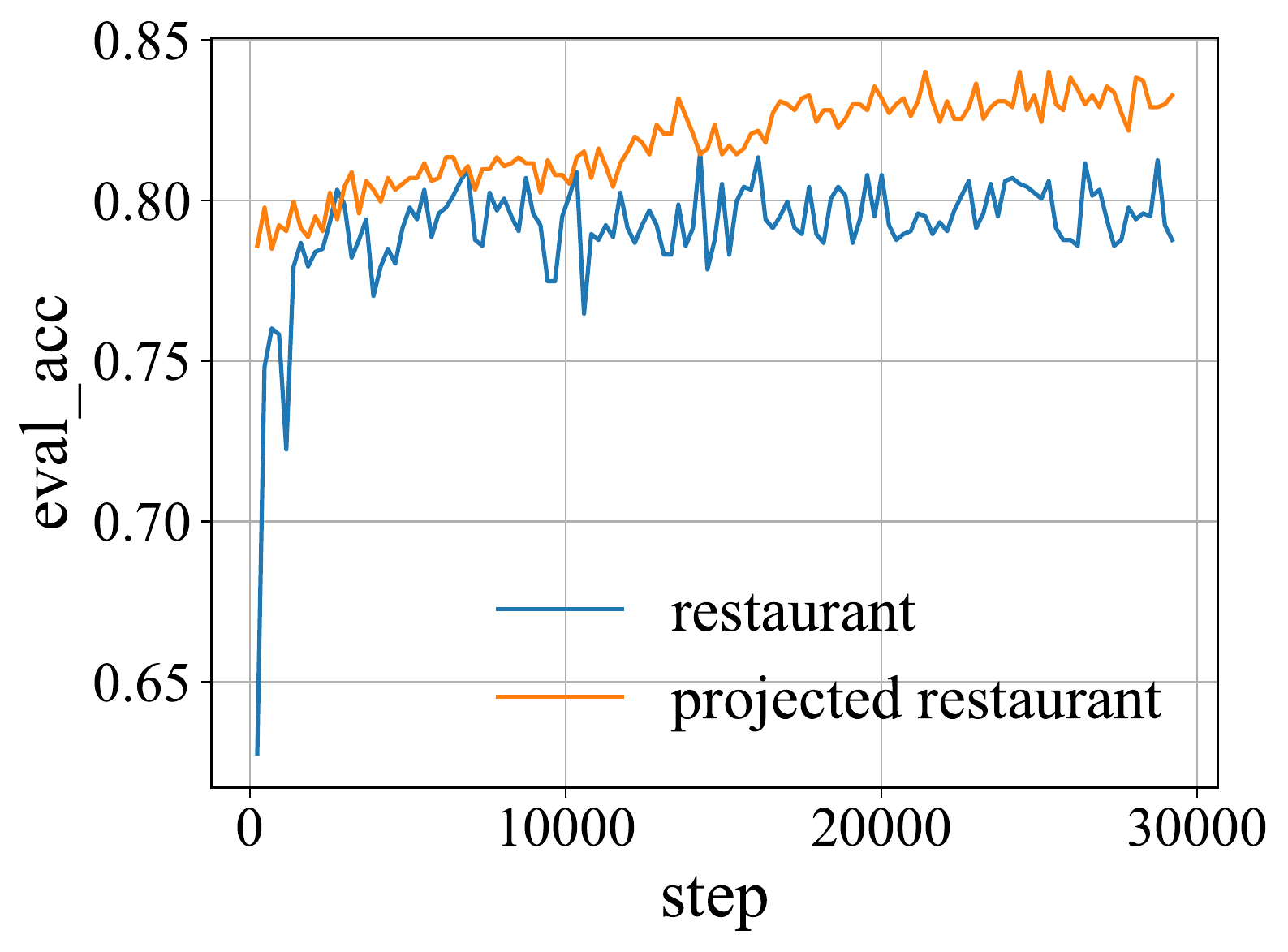}
    }
    
    \caption{The (\textcolor{blue}{\textemdash}) represents vanilla PT on  \RoBERTaBASE. As for (\textcolor{orange}{\textemdash}), it utilizes projected prompts from \BertBASE as initializations to conduct PT on \RoBERTaBASE. The projected prompts respectively come from two different Task Tuning projectors: \texttt{[Without LayerNorm]} and \texttt{[With LayerNorm]}.}
    \label{fig:appendix_no_layernorm_transfer_model_coverage_all}
\end{figure*}

\subsection{More Zero-shot Transfer Performance}
\label{ssec:appendix_zero-shot_transfer_performance_on_various_plms}
In \cref{ssec:zero-shot_transferability_cross-model}, we showed the zero-shot transfer performance of various projector-learning methods in the setting of transferring from \RoBERTaLARGE to \TXXL. We explore more cross-model transfer settings here, which are transferring between various PLMs in different scales and heterogeneous frameworks, including from \BertBASE to \RoBERTaBASE, from \RoBERTaBASE to \RoBERTaLARGE, and from \TBASE to \TXXL. We can find that the results in Table~\ref{table:appendix_various_plms_to_various_plms} are all consistent with \cref{ssec:zero-shot_transferability_cross-model}.

\begin{table*}[!ht]
\begin{center} 
\begin{adjustbox}{max width=0.98\linewidth}
{
\setlength\tabcolsep{0.25em}

\begin{tabular}{cl|cccccc|ccc|cc|cc}
\toprule
\multicolumn{2}{c|}{\multirow{2}{*}{\textbf{Method}}} & \multicolumn{6}{c|}{\textbf{SA}} & \multicolumn{3}{c|}{\textbf{NLI}} & \multicolumn{2}{c|}{\textbf{EJ}} & \multicolumn{2}{c}{\textbf{PI}} \\ \cline{3-15} 
\multicolumn{2}{c|}{} & \multicolumn{1}{c}{IMDB} & \multicolumn{1}{c}{SST-2} & \multicolumn{1}{c}{laptop} & \multicolumn{1}{c}{restaurant} & \multicolumn{1}{c}{Movie} & \multicolumn{1}{c|}{Tweet} & \multicolumn{1}{c}{MNLI} & \multicolumn{1}{c}{QNLI} & \multicolumn{1}{c|}{SNLI} & \multicolumn{1}{c}{deontology} & \multicolumn{1}{c|}{justice} & \multicolumn{1}{c}{QQP} & \multicolumn{1}{c}{MRPC} \\ \midrule
\multicolumn{2}{c|}{PT on \RoBERTaBASE} & 89.9 & 93.8 & 77.3 & 80.7 & 79.2 & 74.5 & 80.6 & 90.5 & 88.5 & 72.9 & 70.0 & 86.9 & 83.9 \\
\midrule
\multicolumn{15}{c}{\texttt{[Without LayerNorm]}} \\
\midrule
\multicolumn{2}{l|}{Task Tuning (IMDB, laptop)} & \textbf{86.5} & \textbf{84.9} & \textbf{73.4} & \textbf{75.3} & \textbf{76.6} & 47.7 & 31.8 & 52.0 & 32.9 & 50.3 & {50.0} & 37.6 & 67.5\\
\multicolumn{2}{l|}{Task Tuning (MNLI)} & 66.6 & 70.4 & 53.0 & 43.8 & 57.8 & 47.9 & \textbf{82.4} & \textbf{74.9} & \textbf{78.1} & {50.4} & 49.9 & {45.3} & {70.1} \\ 
\midrule
\multicolumn{15}{c}{\texttt{[With LayerNorm]}} \\
\midrule
\multicolumn{2}{l|}{Task Tuning (IMDB, laptop)} & \textbf{88.2} & \textbf{82.2} & \textbf{76.3} & \textbf{77.9} & \textbf{73.4} & 43.6 & 32.0 & 47.9 & 32.8 & 49.8 & 49.4 & {50.2} & 47.7\\
\multicolumn{2}{l|}{Task Tuning (MNLI)} & 50.9 & 52.0 & 11.9 & 13.1 & 45.8 & 18.2 & \textbf{80.0} & \textbf{74.9} & \textbf{80.0} & {50.4} & {49.9} & 36.8 & {68.1} \\ 
\bottomrule
\end{tabular}
}
\end{adjustbox}
\caption{We find that the zero-shot performances of prompts projected by two Task Tuning projectors (\texttt{[With LayerNorm]} and \texttt{[Without LayerNorm]}) are close. \textbf{Bold} represents non-trivial performance.}
\label{table:appendix_bertbase_to_robertabase}
\end{center} 
\end{table*}

%\textbf{Bold} represents non-trivial performance.

\subsection{Technical Details of \TPTmodel} %(Transfer with Initialization)}
\label{ssec:appendix_technical_details_of_tptmodel}
In \cref{ssec:transferring_with_initialization_cross-model}, we demonstrate cross-model transferrable prompt tuning (\TPTmodel) can well improve performance and reduce training time. 

However, when we apply \TPTmodel to more PLMs, we find that the projected prompts may have quite different $L_{2}$ norm values with the original prompts, especially for the small-scale PLMs (e.g., from \BertBASE to \RoBERTaBASE). Specifically, we obtain the projected prompts with the trained Task Tuning projector, and find that the projected prompts are hard to optimize in some tasks as shown in Figure~\ref{fig:appendix_no_layernorm_transfer_model_coverage_all} \texttt{[Without LayerNorm]}. Thus, we attempt to add the layer normalization operation~\citep{ba2016layer} \texttt{LayerNorm} into the projectors to regularize the norm of the projected prompt as follows:
\begin{equation}
\label{eq:appendix_transfer_projector}
\begin{aligned}
    \mathbf{\tilde{P}}^{s} = \texttt{LayerNorm}(\mathbf{Proj}(\mathbf{P}^{s})).
\end{aligned}
\end{equation}
By the \texttt{LayerNorm}, the projected prompts can work well on \TPTmodel and  achieve better performance and speedup as shown in Figure \ref{fig:appendix_no_layernorm_transfer_model_coverage_all} \texttt{[With LayerNorm]}. Interestingly, although prompts projected by the projectors \texttt{[Without LayerNorm]} are hard to be trained in \TPTmodel, they can achieve similar zero-shot transfer performance with the prompts projected by the projectors \texttt{[With LayerNorm]} in Table~\ref{table:appendix_bertbase_to_robertabase}.

\begin{table}[!t]
\begin{center} 
\small
\scalebox{0.925}
{{
\setlength\tabcolsep{0.4em}
\begin{tabular}{l|c|c|c}
\toprule
{\textbf{Metric}} & \multicolumn{1}{c|}{\makecell{Same\\Tasks}} & \multicolumn{1}{c|}{\makecell{Same-type\\Tasks}}   & \multicolumn{1}{c}{\makecell{Different-type\\Tasks}} \\
\midrule
\multicolumn{4}{c}{\RoBERTaLARGE} \\
\midrule
{$\mathrm{E}_{\mathrm{concat}}$ } & \multicolumn{1}{r|}{9.4} & \multicolumn{1}{r|}{9.4} & \multicolumn{1}{r}{6.8}\\  
% \midrule
{$\mathrm{E}_{\mathrm{average}}$} & \multicolumn{1}{r|}{41.6} & \multicolumn{1}{r|}{41.4} & \multicolumn{1}{r}{37.6}\\  
% \midrule
{$\mathrm{C}_{\mathrm{concat}}$ } & \multicolumn{1}{r|}{47.6} & \multicolumn{1}{r|}{45.3} & \multicolumn{1}{r}{31.7}\\ 
% \midrule
{$\mathrm{C}_{\mathrm{average}}$} & \multicolumn{1}{r|}{1.7} & \multicolumn{1}{r|}{1.3} & \multicolumn{1}{r}{1.1}\\ 
%\midrule
% \midrule
{$\mathrm{ON}$ (Bottom 3)} & \multicolumn{1}{r|}{42.8} & \multicolumn{1}{r|}{43.3} & \multicolumn{1}{r}{39.1}\\ 
{$\mathrm{ON}$ (Top 3)} & \multicolumn{1}{r|}{39.4} & \multicolumn{1}{r|}{28.2} & \multicolumn{1}{r}{21.4}\\ 
{$\mathrm{ON}$ (All 24) } & \multicolumn{1}{r|}{40.0} & \multicolumn{1}{r|}{35.8} & \multicolumn{1}{r}{29.6}\\ 
\midrule
\multicolumn{4}{c}{\TXXL(Decoder Module)} \\
\midrule
{$\mathrm{E}_{\mathrm{concat}}$ } & \multicolumn{1}{r|}{0.5} & \multicolumn{1}{r|}{0.5} & \multicolumn{1}{r}{0.3}\\  
% \midrule
{$\mathrm{E}_{\mathrm{average}}$} & \multicolumn{1}{r|}{4.0} & \multicolumn{1}{r|}{5.1} & \multicolumn{1}{r}{3.4}\\  
% \midrule
{$\mathrm{C}_{\mathrm{concat}}$ } & \multicolumn{1}{r|}{29.4} & \multicolumn{1}{r|}{2.8} & \multicolumn{1}{r}{2.4}\\ 
% \midrule
{$\mathrm{C}_{\mathrm{average}}$} & \multicolumn{1}{r|}{4.0} & \multicolumn{1}{r|}{2.6} & \multicolumn{1}{r}{2.1}\\ 
%\midrule
% \midrule
{$\mathrm{ON}$ (Bottom 3) } & \multicolumn{1}{r|}{80.3} & \multicolumn{1}{r|}{75.4} & \multicolumn{1}{r}{76.3}\\ 
{$\mathrm{ON}$ (Top 3)} & \multicolumn{1}{r|}{62.0} & \multicolumn{1}{r|}{52.7} & \multicolumn{1}{r}{46.1}\\ 
{$\mathrm{ON}$ (All 24) } & \multicolumn{1}{r|}{60.8} & \multicolumn{1}{r|}{54.0} & \multicolumn{1}{r}{49.2}\\ 
\bottomrule
\end{tabular}
}}
\caption{The average values (\%) of the $5$ similarity metrics for prompt pairs within the same task (trained with $3$ different random seeds) and between different tasks (of the same type and different types) on \RoBERTaLARGE and \TXXL.}
\label{table:appendix_similiarty_metrics}
\end{center} 
\end{table}

\section{Transferability Indicator}
\label{sec:appendix_transferability_indicator}

\subsection{Effectiveness of Similarity Metrics}
\label{ssec:appendix_effectiveness_of_similarity_metrics}
We categorize all prompts into three groups: same tasks (prompts trained with different seeds on the same dataset), same-type tasks, and different-type tasks. Table~\ref{table:appendix_similiarty_metrics} shows that all the similarity metrics successfully distinguish task types.

\begin{table}[!t]
\begin{center} 
% \small
\scalebox{0.73}
{
{
\setlength\tabcolsep{0.3em}
\begin{tabular}{l|rrrrrr|r}
\toprule
{\textbf{Metric}} & \multicolumn{1}{c}{SA} & \multicolumn{1}{c}{NLI} & \multicolumn{1}{c}{EJ} & \multicolumn{1}{c}{PI} & \multicolumn{1}{c}{QA} & \multicolumn{1}{c|}{SUM} & \multicolumn{1}{c}{All} \\
\midrule
\multicolumn{8}{c}{\TSMALL(Decoder Module)} \\
\midrule
{$\mathrm{E}_{\mathrm{concat}}$ } & \multicolumn{1}{r}{10.1} & \multicolumn{1}{r}{19.6} & \multicolumn{1}{r}{31.3} & \multicolumn{1}{r}{5.3} & \multicolumn{1}{r}{27.3} & \multicolumn{1}{r|}{38.0} & \multicolumn{1}{r}{21.9}\\  
{$\mathrm{E}_{\mathrm{average}}$} & \multicolumn{1}{r}{-6.8} & \multicolumn{1}{r}{-28.0} & \multicolumn{1}{r}{18.7} & \multicolumn{1}{r}{-2.6} & \multicolumn{1}{r}{29.1} & \multicolumn{1}{r|}{42.9} & \multicolumn{1}{r}{8.9}\\  
{$\mathrm{C}_{\mathrm{concat}}$ } & \multicolumn{1}{r}{34.6} & \multicolumn{1}{r}{63.6} & \multicolumn{1}{r}{26.6} & \multicolumn{1}{r}{\textbf{19.3}} & \multicolumn{1}{r}{-2.1} & \multicolumn{1}{r|}{12.5} & \multicolumn{1}{r}{25.7}\\ 
{$\mathrm{C}_{\mathrm{average}}$} & \multicolumn{1}{r}{\textbf{64.3}} & \multicolumn{1}{r}{65.1} & \multicolumn{1}{r}{30.7} & \multicolumn{1}{r}{15.7} & \multicolumn{1}{r}{27.7} & \multicolumn{1}{r|}{19.2} & \multicolumn{1}{r}{37.1}\\ 
\midrule
{$\mathrm{ON}$ (Bottom 3) } & \multicolumn{1}{r}{32.9} & \multicolumn{1}{r}{72.6} & \multicolumn{1}{r}{41.8} & \multicolumn{1}{r}{14.2} & \multicolumn{1}{r}{45.5} & \multicolumn{1}{r|}{52.8} & \multicolumn{1}{r}{43.3}\\ 
{$\mathrm{ON}$ (Top 3) } & \multicolumn{1}{r}{50.6} & \multicolumn{1}{r}{74.8} & \multicolumn{1}{r}{\textbf{51.4}} & \multicolumn{1}{r}{2.6} & \multicolumn{1}{r}{\textbf{60.3}} & \multicolumn{1}{r|}{\textbf{78.8}} & \multicolumn{1}{r}{\textbf{52.5}}\\ 
{$\mathrm{ON}$ (All 24) } & \multicolumn{1}{r}{44.8} & \multicolumn{1}{r}{\textbf{79.7}} & \multicolumn{1}{r}{44.5} & \multicolumn{1}{r}{6.3} & \multicolumn{1}{r}{59.7} & \multicolumn{1}{r|}{67.9} & \multicolumn{1}{r}{50.5}\\ 
\midrule
\multicolumn{8}{c}{\TBASE(Decoder Module)} \\
\midrule
{$\mathrm{E}_{\mathrm{concat}}$ } & \multicolumn{1}{r}{55.2} & \multicolumn{1}{r}{-17.0} & \multicolumn{1}{r}{10.2} & \multicolumn{1}{r}{21.5} & \multicolumn{1}{r}{5.9} & \multicolumn{1}{r|}{-1.1} & \multicolumn{1}{r}{20.8}\\  
{$\mathrm{E}_{\mathrm{average}}$} & \multicolumn{1}{r}{53.4} & \multicolumn{1}{r}{-42.3} & \multicolumn{1}{r}{-10.7} & \multicolumn{1}{r}{7.5} & \multicolumn{1}{r}{-27.7} & \multicolumn{1}{r|}{-10.8} & \multicolumn{1}{r}{9.0}\\  
{$\mathrm{C}_{\mathrm{concat}}$ } & \multicolumn{1}{r}{\textbf{57.2}} & \multicolumn{1}{r}{25.2} & \multicolumn{1}{r}{35.1} & \multicolumn{1}{r}{37.0} & \multicolumn{1}{r}{30.2} & \multicolumn{1}{r|}{-20.5} & \multicolumn{1}{r}{28.4}\\ 
{$\mathrm{C}_{\mathrm{average}}$} & \multicolumn{1}{r}{47.6} & \multicolumn{1}{r}{\textbf{70.0}} & \multicolumn{1}{r}{30.4} & \multicolumn{1}{r}{\textbf{48.0}} & \multicolumn{1}{r}{34.9} & \multicolumn{1}{r|}{16.8} & \multicolumn{1}{r}{42.4}\\ 
\midrule
{$\mathrm{ON}$ (Bottom 3) } & \multicolumn{1}{r}{34.7} & \multicolumn{1}{r}{29.8} & \multicolumn{1}{r}{40.8} & \multicolumn{1}{r}{16.9} & \multicolumn{1}{r}{24.2} & \multicolumn{1}{r|}{72.2} & \multicolumn{1}{r}{36.0}\\ 
{$\mathrm{ON}$ (Top 3) } & \multicolumn{1}{r}{53.8} & \multicolumn{1}{r}{24.3} & \multicolumn{1}{r}{\textbf{50.6}} & \multicolumn{1}{r}{46.1} & \multicolumn{1}{r}{54.7} & \multicolumn{1}{r|}{\textbf{79.1}} & \multicolumn{1}{r}{\textbf{49.1}}\\ 
{$\mathrm{ON}$ (All 24) } & \multicolumn{1}{r}{46.1} & \multicolumn{1}{r}{25.0} & \multicolumn{1}{r}{42.6} & \multicolumn{1}{r}{39.7} & \multicolumn{1}{r}{\textbf{56.7}} & \multicolumn{1}{r|}{72.3} & \multicolumn{1}{r}{43.4}\\ 
\midrule
\multicolumn{8}{c}{\TXXL(Decoder Module)} \\
\midrule
{$\mathrm{E}_{\mathrm{concat}}$ } & \multicolumn{1}{r}{40.8} & \multicolumn{1}{r}{-13.4} & \multicolumn{1}{r}{19.3} & \multicolumn{1}{r}{11.4} & \multicolumn{1}{r}{-4.3} & \multicolumn{1}{r|}{-19.5} & \multicolumn{1}{r}{12.9}\\ 
{$\mathrm{E}_{\mathrm{average}}$} & \multicolumn{1}{r}{32.2} & \multicolumn{1}{r}{-42.6} & \multicolumn{1}{r}{9.7} & \multicolumn{1}{r}{-2.0} & \multicolumn{1}{r}{-27.7} & \multicolumn{1}{r|}{-34.0} & \multicolumn{1}{r}{-2.5}\\ 
{$\mathrm{C}_{\mathrm{concat}}$ } & \multicolumn{1}{r}{21.4} & \multicolumn{1}{r}{40.9} & \multicolumn{1}{r}{\textbf{42.6}} & \multicolumn{1}{r}{24.6} & \multicolumn{1}{r}{30.2} & \multicolumn{1}{r|}{45.6} & \multicolumn{1}{r}{31.6}\\ 
{$\mathrm{C}_{\mathrm{average}}$} & \multicolumn{1}{r}{23.3} & \multicolumn{1}{r}{\textbf{44.8}} & \multicolumn{1}{r}{33.3} & \multicolumn{1}{r}{29.3} & \multicolumn{1}{r}{34.9} & \multicolumn{1}{r|}{\textbf{49.9}} & \multicolumn{1}{r}{33.5}\\ 
\midrule
{$\mathrm{ON}$ (Bottom 3) } & \multicolumn{1}{r}{9.1} & \multicolumn{1}{r}{20.7} & \multicolumn{1}{r}{14.8} & \multicolumn{1}{r}{18.3} & \multicolumn{1}{r}{24.2} & \multicolumn{1}{r|}{-9.9} & \multicolumn{1}{r}{12.4}\\ 
{$\mathrm{ON}$ (Top 3) } & \multicolumn{1}{r}{\textbf{42.7}} & \multicolumn{1}{r}{33.6} & \multicolumn{1}{r}{39.1} & \multicolumn{1}{r}{30.3} & \multicolumn{1}{r}{54.7} & \multicolumn{1}{r|}{11.1} & \multicolumn{1}{r}{\textbf{36.9}}\\ 
{$\mathrm{ON}$ (All 24) } & \multicolumn{1}{r}{31.0} & \multicolumn{1}{r}{23.6} & \multicolumn{1}{r}{37.7} & \multicolumn{1}{r}{\textbf{34.2}} & \multicolumn{1}{r}{\textbf{56.7}} & \multicolumn{1}{r|}{15.4} & \multicolumn{1}{r}{32.0}\\ 
% \midrule
\midrule
% \multicolumn{8}{c}{\TXXL (\textbf{I}) (Decoder Module)} \\
% \midrule
{$\mathrm{ON}_{\mathbf{I}}$ (Bottom 3) } & \multicolumn{1}{c}{- -} & \multicolumn{1}{c}{- -} & \multicolumn{1}{c}{- -} & \multicolumn{1}{c}- -{} & \multicolumn{1}{c}{- -} & \multicolumn{1}{c|}{- -} & \multicolumn{1}{c}{\textbf{25.3}}\\ 
{$\mathrm{ON}_{\mathbf{I}}$ (Top 3) } & \multicolumn{1}{c}{- -} & \multicolumn{1}{c}{- -} & \multicolumn{1}{c}{- -} & \multicolumn{1}{c}{- -} & \multicolumn{1}{c}{- -} & \multicolumn{1}{c|}{- -} & \multicolumn{1}{c}{\textbf{46.3}}\\ 
{$\mathrm{ON}_{\mathbf{I}}$ (All 24) } & \multicolumn{1}{c}{- -} & \multicolumn{1}{c}{- -} & \multicolumn{1}{c}{- -} & \multicolumn{1}{c}{- -} & \multicolumn{1}{c}{- -} & \multicolumn{1}{c|}{- -} & \multicolumn{1}{c}{\textbf{40.0}}\\ 
\bottomrule
\end{tabular}}}
\caption{Spearman’s rank correlation scores (\%) between various similarity metrics and zero-shot transfer performance of soft prompts for various scales of T5 and $\mathrm{ON}_{\mathbf{I}}$ as introduced in \cref{ssec:appendix_plms_high_redundancy}.}
\label{table:appendix_similiarty_metrics_t5}
\end{center} 
\end{table}

\begin{table}[!t]
\begin{center} 
% \small
\scalebox{0.8}
{
{
\setlength\tabcolsep{0.6em}
\begin{tabular}{l|rrrr|r}
\toprule
{\textbf{Metric}} & \multicolumn{1}{c}{SA} & \multicolumn{1}{c}{NLI} & \multicolumn{1}{c}{EJ} & \multicolumn{1}{c|}{PI} & \multicolumn{1}{c}{All} \\
\midrule
\multicolumn{6}{c}{\RoBERTaBASE} \\
\midrule
{$\mathrm{E}_{\mathrm{concat}}$ } & \multicolumn{1}{r}{31.1} & \multicolumn{1}{r}{-5.9} & \multicolumn{1}{r}{30.5} & \multicolumn{1}{r|}{16.2} & \multicolumn{1}{r}{20.2} \\  
{$\mathrm{E}_{\mathrm{average}}$} & \multicolumn{1}{r}{17.2} & \multicolumn{1}{r}{-52.4} & \multicolumn{1}{r}{12.1} & \multicolumn{1}{r|}{-13.5} & \multicolumn{1}{r}{-4.4} \\  
{$\mathrm{C}_{\mathrm{concat}}$ } & \multicolumn{1}{r}{51.6} & \multicolumn{1}{r}{8.8} & \multicolumn{1}{r}{38.5} & \multicolumn{1}{r|}{29.7} & \multicolumn{1}{r}{36.3} \\ 
{$\mathrm{C}_{\mathrm{average}}$} & \multicolumn{1}{r}{65.8} & \multicolumn{1}{r}{55.9} & \multicolumn{1}{r}{26.1} & \multicolumn{1}{r|}{28.9} & \multicolumn{1}{r}{51.7} \\ 
\midrule
{$\mathrm{ON}$ (Bottom 3) } & \multicolumn{1}{r}{56.2} & \multicolumn{1}{r}{64.3} & \multicolumn{1}{r}{17.9} & \multicolumn{1}{r|}{21.2} & \multicolumn{1}{r}{46.8} \\ 
{$\mathrm{ON}$ (Top 3) } & \multicolumn{1}{r}{\textbf{77.9}} & \multicolumn{1}{r}{\textbf{74.2}} & \multicolumn{1}{r}{\textbf{43.4}} & \multicolumn{1}{r|}{\textbf{32.7}} & \multicolumn{1}{r}{\textbf{64.8}} \\ 
{$\mathrm{ON}$ (All 24) } & \multicolumn{1}{r}{71.2} & \multicolumn{1}{r}{70.5} & \multicolumn{1}{r}{33.6} & \multicolumn{1}{r|}{25.0} & \multicolumn{1}{r}{58.1} \\ 
\midrule
\multicolumn{6}{c}{\RoBERTaLARGE} \\
\midrule
{$\mathrm{E}_{\mathrm{concat}}$ } & \multicolumn{1}{r}{42.5} & \multicolumn{1}{r}{-16.3} & \multicolumn{1}{r}{21.4} & \multicolumn{1}{r|}{22.8} & \multicolumn{1}{r}{22.6} \\  
{$\mathrm{E}_{\mathrm{average}}$} & \multicolumn{1}{r}{34.5} & \multicolumn{1}{r}{-55.1} & \multicolumn{1}{r}{-5.8} & \multicolumn{1}{r|}{3.6} & \multicolumn{1}{r}{2.8} \\  
{$\mathrm{C}_{\mathrm{concat}}$ } & \multicolumn{1}{r}{44.5} & \multicolumn{1}{r}{-11.7} & \multicolumn{1}{r}{23.6} & \multicolumn{1}{r|}{22.0} & \multicolumn{1}{r}{24.8} \\ 
{$\mathrm{C}_{\mathrm{average}}$} & \multicolumn{1}{r}{38.2} & \multicolumn{1}{r}{77.1} & \multicolumn{1}{r}{12.4} & \multicolumn{1}{r|}{\textbf{47.8}} & \multicolumn{1}{r}{44.7} \\ 
\midrule
{$\mathrm{ON}$ (Bottom 3) } & \multicolumn{1}{r}{32.0} & \multicolumn{1}{r}{34.8} & \multicolumn{1}{r}{\textbf{44.5}} & \multicolumn{1}{r|}{30.3} & \multicolumn{1}{r}{34.3} \\ 
{$\mathrm{ON}$ (Top 3) } & \multicolumn{1}{r}{\textbf{70.9}} & \multicolumn{1}{r}{\textbf{45.6}} & \multicolumn{1}{r}{13.5} & \multicolumn{1}{r|}{28.9} & \multicolumn{1}{r}{\textbf{49.7}} \\ 
{$\mathrm{ON}$ (All 24) } & \multicolumn{1}{r}{62.7} & \multicolumn{1}{r}{40.6} & \multicolumn{1}{r}{16.0} & \multicolumn{1}{r|}{31.1} & \multicolumn{1}{r}{45.6} \\ 
\bottomrule
\end{tabular}}}
\caption{Spearman’s rank correlation scores (\%) between various similarity metrics and zero-shot transfer performance of soft prompts for various scales of RoBERTa.}
\label{table:appendix_similiarty_metrics_roberta}
\end{center} 
\end{table}

\subsection{Correlation Between Prompt Transferability and Prompt Similarity}
\label{ssec:appendix_correlation_between_prompt_transferability_and_prompt_similarity}
In \cref{ssec:cross-task_prompt_similarity}, we provide the overall averaged Spearman’s rank correlation scores (\%) between various similarity metrics and zero-shot transfer performance of soft prompts for \RoBERTaLARGE and \TXXL. 

Here, we further show Spearman’s rank correlation scores grouped by the task types on more PLMs. The results are shown in Table~\ref{table:appendix_similiarty_metrics_t5} and Table~\ref{table:appendix_similiarty_metrics_roberta}.

\subsection{PLMs' Redundancy Influence Indicators}
\label{ssec:appendix_plms_high_redundancy}
From Table \ref{table:appendix_similiarty_metrics_t5}, we find that the correlation between prompt transferability and prompt similarity will drop with the increase of PLM size. We guess that this phenomena may result from PLMs' high redundancy~\citep{Aghajanyan2021IntrinsicDE}. 

To try to overcome this, we simultaneously utilize the prompts trained with three random seeds on the same dataset and take their intersection of activation states as the activated neurons into the similarity ($\mathrm{ON}$) computation. This similarity is called $\mathrm{ON}_{\mathbf{I}}$. By using it, the correlation score of ON can significantly raise as shown in Table~\ref{table:appendix_similiarty_metrics_t5}.

% \begin{table}[!htbp]
% \begin{center} 
% \scalebox{1.0}
% {
% {
% \setlength\tabcolsep{0.16em}
% \begin{tabular}{l|r|r}
% \toprule
%  \multicolumn{1}{l|}{\textbf{Correlation~(\%)}} & \multicolumn{2}{c}{\textbf{Model}} \\
%  \midrule
%  {\textbf{Metrics}} & \multicolumn{1}{c|}{\TXXL} & \multicolumn{1}{c}{\TXXL(I)}\\
% \midrule
% $\mathrm{E}_{\mathrm{concat}}$  & \multicolumn{1}{c|}{12.9} & \multicolumn{1}{c}{- -}  \\
% $\mathrm{E}_{\mathrm{average}}$ & \multicolumn{1}{c|}{-2.5} & \multicolumn{1}{c}{- -}  \\
% $\mathrm{C}_{\mathrm{concat}}$ & \multicolumn{1}{c|}{31.6} & \multicolumn{1}{c}{- -}  \\
% $\mathrm{C}_{\mathrm{average}}$ & \multicolumn{1}{c|}{33.5} & \multicolumn{1}{c}{- -}  \\
% \midrule
% {{$\mathrm{ON}$ (The bottom 3 layers)}} & \multicolumn{1}{c|}{12.4} & \multicolumn{1}{c}{\textbf{25.3}} \\
% {$\mathrm{ON}$ (The top 3 layers)} & \multicolumn{1}{c|}{49.7} & \multicolumn{1}{c}{\textbf{46.3}} \\
% {{$\mathrm{ON}$ (All 24 layers)}} & \multicolumn{1}{c|}{32.0} & \multicolumn{1}{c}{\textbf{40.0}} \\
% \bottomrule
% \end{tabular}
% }}
% \caption{\TXXL(I) indicates that we utilize the prompts (trained with 3 different random seeds on the same dataset) and take their intersection of activation states into computation.} 
% \label{table:appendix_prompt_similiarty_measure}
% \end{center} 
% \end{table}

\subsection{Overlapping Rate of Activated Neurons in Different Layers}
\label{ssec:appendix_overlapping_percentages_of_activated_neurons}
\begin{figure*}[!htbp]
\centering
\includegraphics[width=0.925\textwidth]{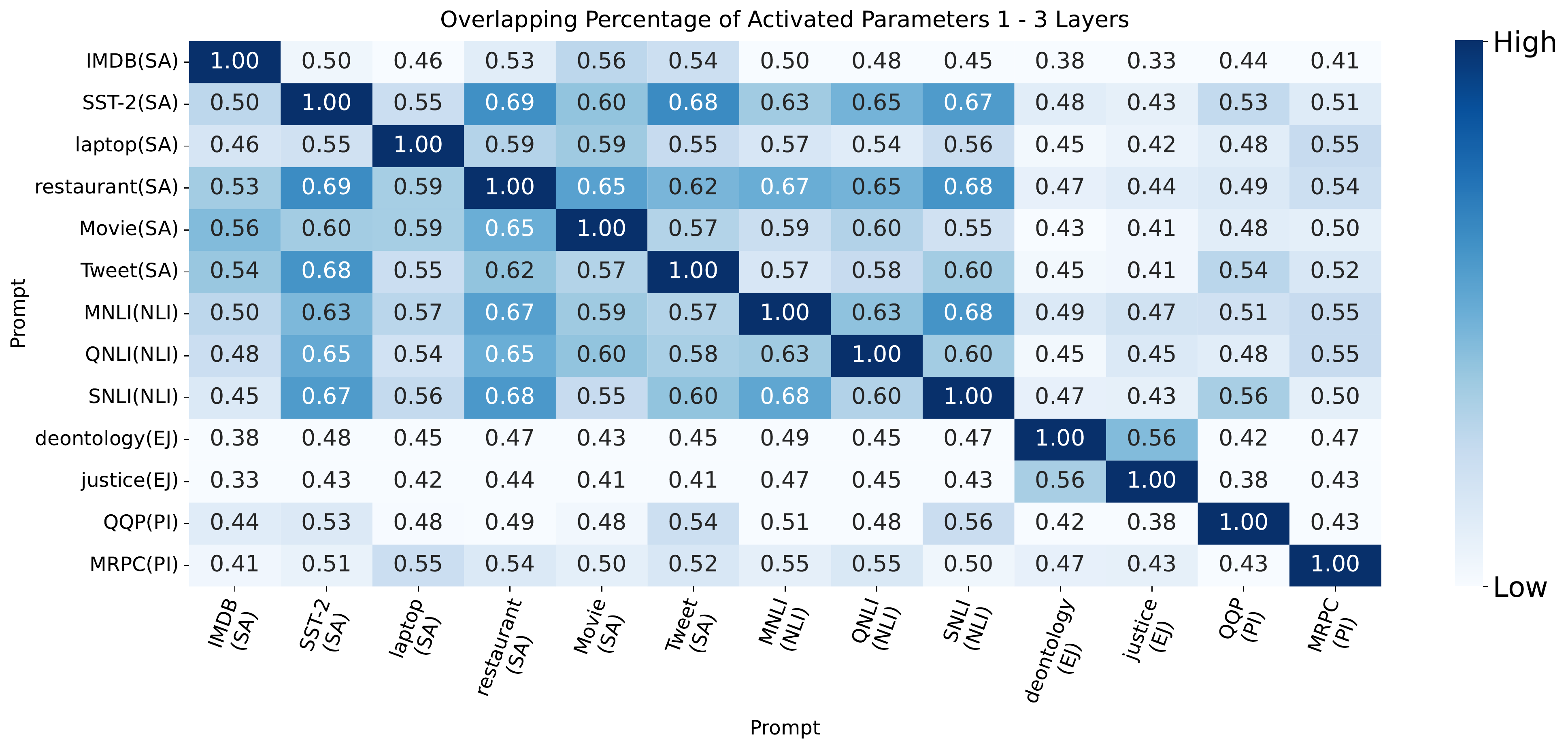}
\caption{ON in 1 - 3 layers of \RoBERTaBASE.}
\label{fig:task_overlap_1-3_layers}
\end{figure*}

\begin{figure*}[!htbp]
\centering
\includegraphics[width=0.925\textwidth]{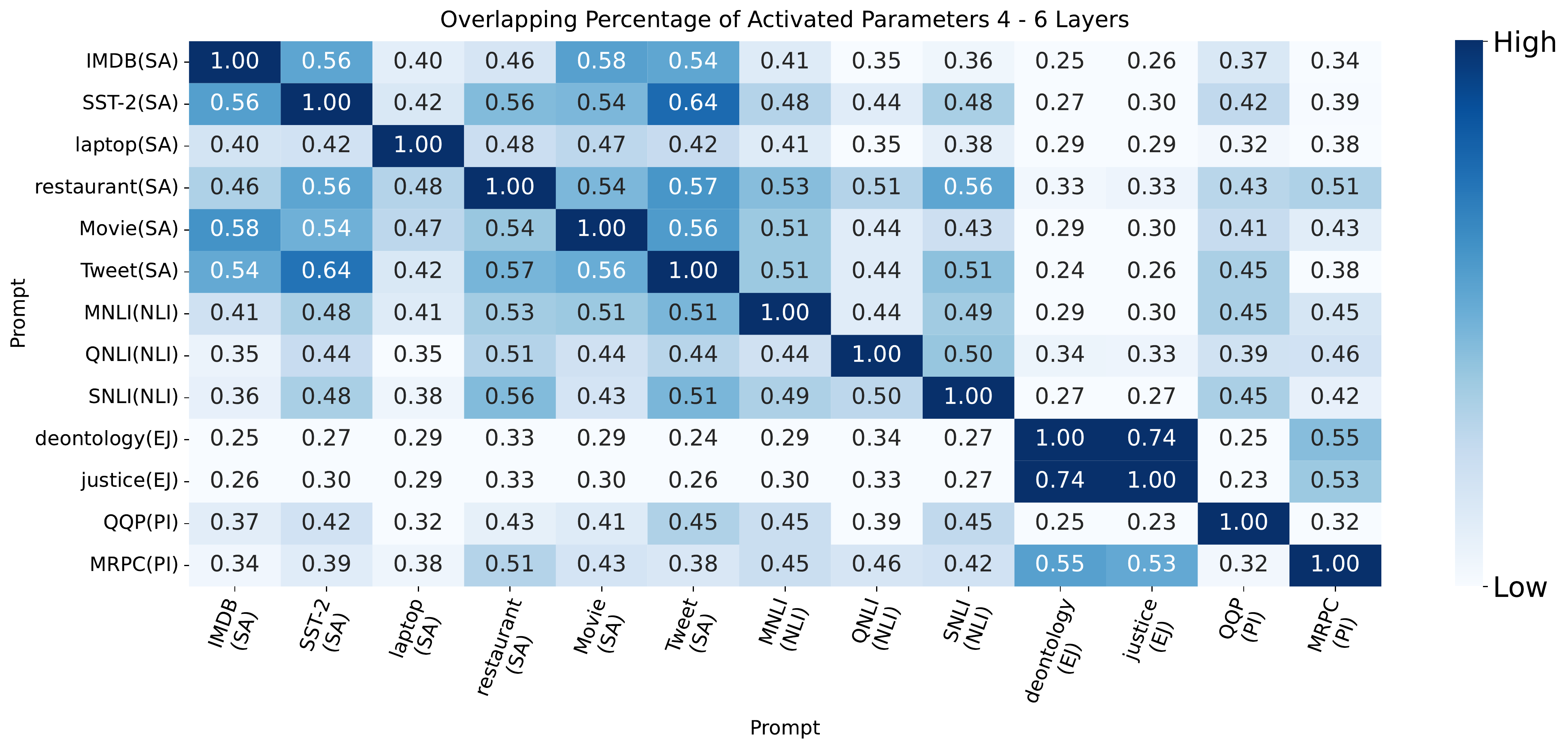}
\caption{ON in 4 - 6 layers of \RoBERTaBASE.}
\label{fig:task_overlap_4-6_layers}
\end{figure*}

\begin{figure*}[!htbp]
\centering
\includegraphics[width=0.925\textwidth]{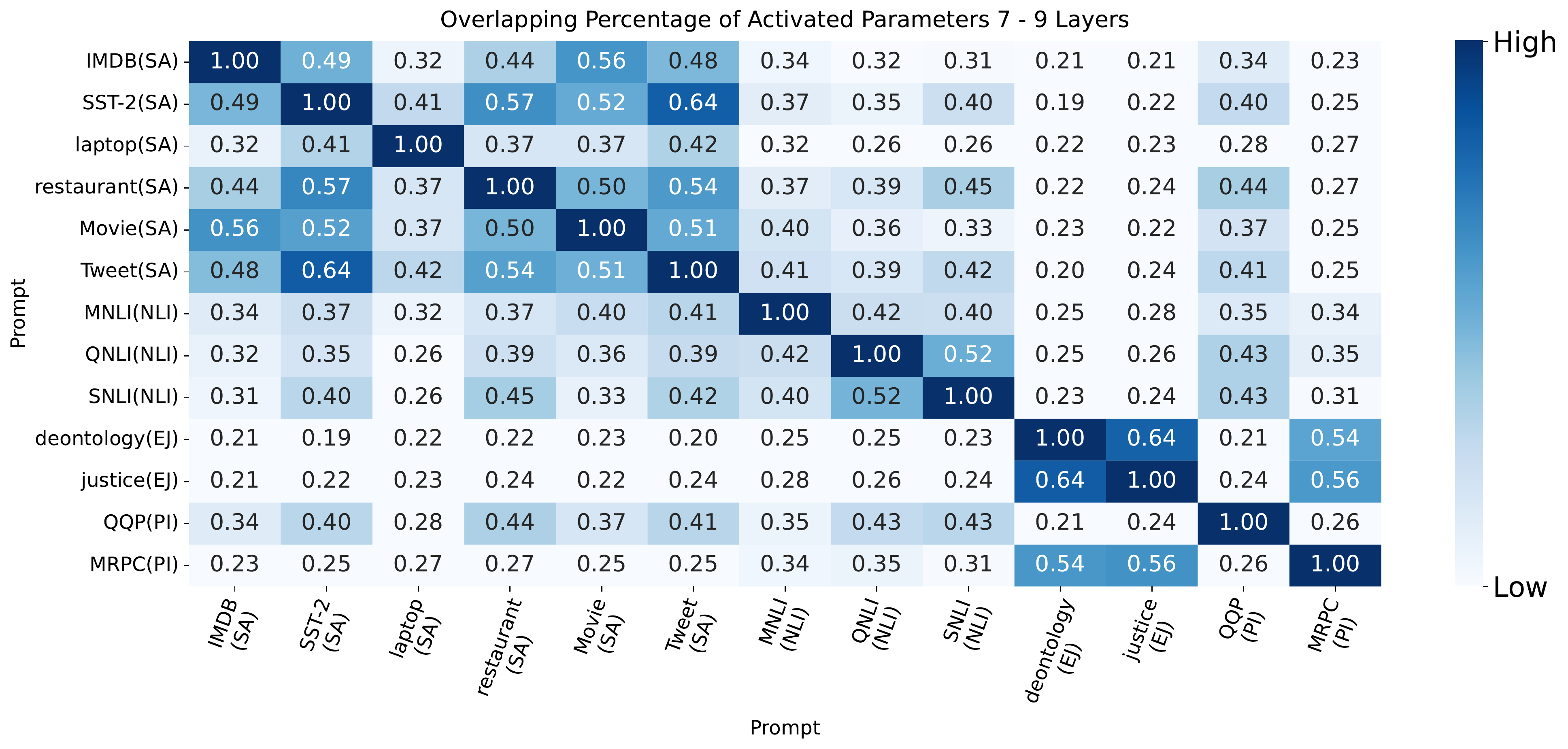}
\caption{ON in 7 - 9 layers of \RoBERTaBASE.}
\label{fig:task_overlap_7-9_layers}
\end{figure*}

\begin{figure*}[!htbp]
\centering
\includegraphics[width=0.925\textwidth]{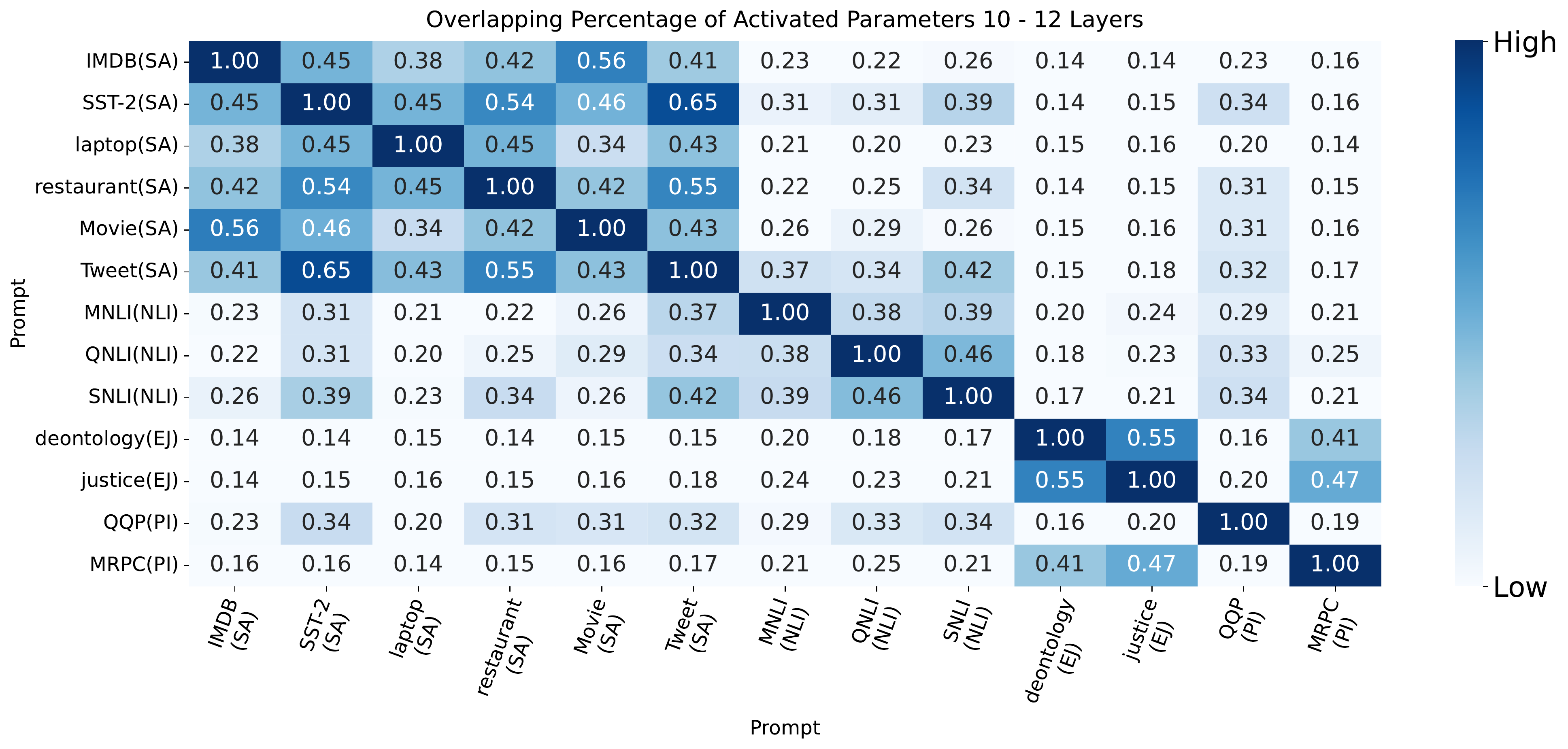}
\caption{ON in 10 - 12 layers of \RoBERTaBASE.}
\label{fig:task_overlap_10-12_layers}
\end{figure*}

\begin{figure*}[!htbp]
\centering
\includegraphics[width=0.925\textwidth]{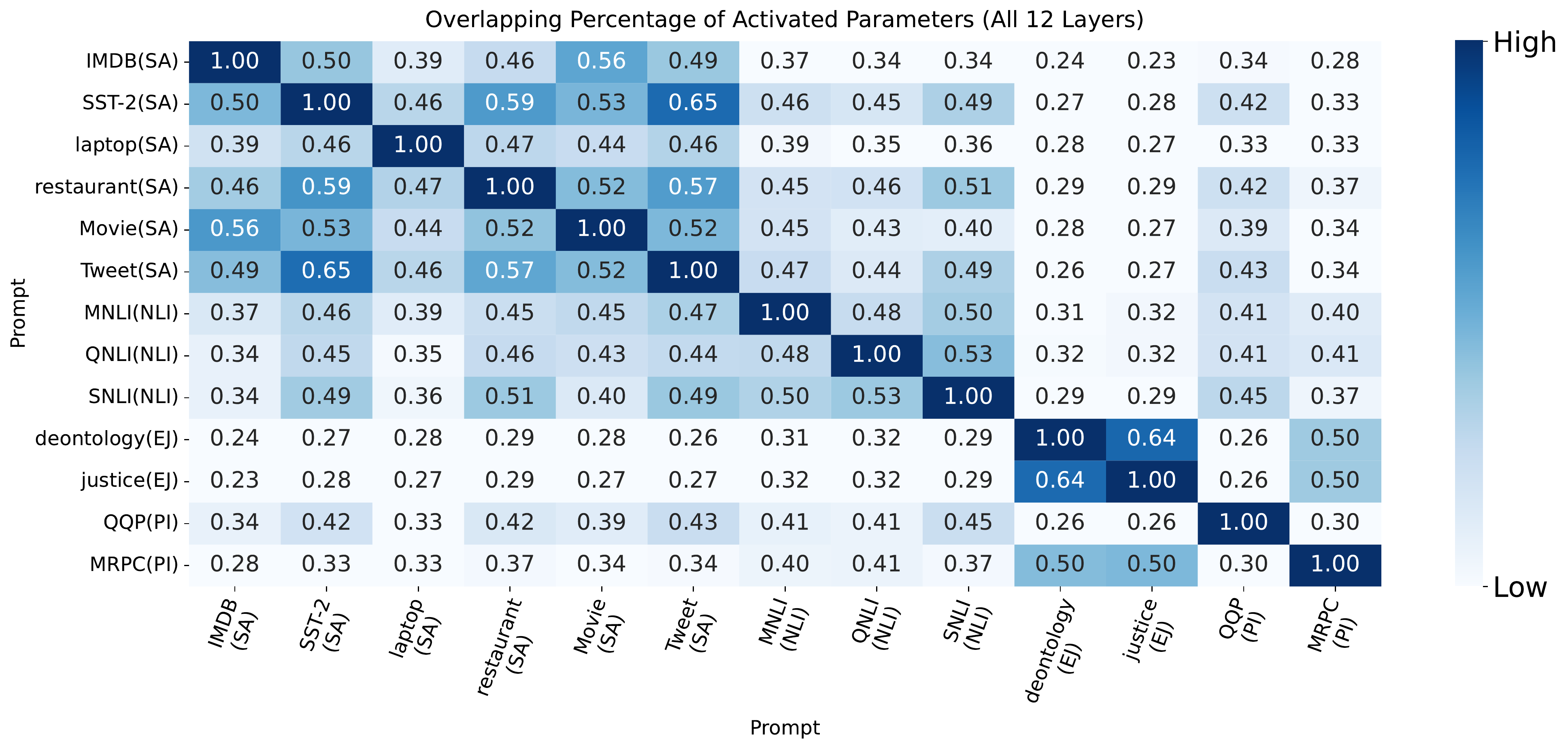}
\caption{ON in all 12 layers of \RoBERTaBASE.}
\label{fig:task_overlap_12_layers}
\end{figure*}

To further understand model stimulation in PLMs, we investigate $\mathrm{ON}$ in different layers of PLMs. Specifically, on \RoBERTaBASE, we measure the similarity between different prompts with activation states of from 1 to 3 layers (Figure~\ref{fig:task_overlap_1-3_layers}), from 4 to 6 layers (Figure~\ref{fig:task_overlap_4-6_layers}), from 7 to 9 layers (Figure~\ref{fig:task_overlap_7-9_layers}), from 10 to 12 layers (Figure~\ref{fig:task_overlap_10-12_layers}), and all 12 layers (Figure~\ref{fig:task_overlap_12_layers}), respectively. 

We find that the activated neurons are common in the bottom layers but tend to be more task-specific in top layers, which is consistent with the findings of previous works~\citep{Liu2019LinguisticKA}.

